\def\ARXIVBUILD{1}
\documentclass[11pt]{article}

\usepackage[
  letterpaper,
  left=0.92in,
  right=0.92in,
  top=0.4in,
  bottom=1.15in,
  headsep=0.16in,
  footskip=0.38in
]{geometry}
\usepackage{times}
\usepackage[T1]{fontenc}
\usepackage{microtype}
\usepackage{graphicx}
\usepackage{amsmath,amssymb,amsthm,mathtools,booktabs}
\usepackage{multirow}
\usepackage[table]{xcolor}
\usepackage{xspace}
\usepackage{pifont}
\usepackage{float}
\usepackage{subcaption}
\usepackage{wrapfig}
\usepackage{longtable}
\usepackage{needspace}
\usepackage[authoryear,round]{natbib}
\usepackage{titlesec}
\usepackage[most]{tcolorbox}
\usepackage{fancyhdr}
\usepackage{url}
\usepackage{hyperref}

\usepackage{amsmath,amsfonts,bm}

\def\eqref#1{equation~\ref{#1}}

\def\1{\bm{1}}

\DeclareMathAlphabet{\mathsfit}{\encodingdefault}{\sfdefault}{m}{sl}
\SetMathAlphabet{\mathsfit}{bold}{\encodingdefault}{\sfdefault}{bx}{n}

\hypersetup{
  colorlinks=true,
  linkcolor=blue!65!black,
  citecolor=blue!65!black,
  urlcolor=blue!70!black,
  pdfauthor={Yifan Wang et al.},
  pdftitle={Model-Aware Data Selection from In-and-Out Information Interplay}
}

\newtheorem{remark}{Remark}
\newtheorem{proposition}{Proposition}
\newtheorem{definition}{Definition}
\newtheorem{theorem}{Theorem}

\newtheorem{corollary}{Corollary}
\theoremstyle{definition}
\newtheorem{assumption}{Assumption}

\definecolor{locushl}{RGB}{219,234,254}
\definecolor{gaingreen}{RGB}{22,163,74}
\definecolor{circlednum}{RGB}{0,0,0}
\definecolor{settinghl}{RGB}{255,244,214}
\definecolor{abstractbg}{RGB}{235,247,252}

\newcommand{\Dearly}{D_{\mathrm{early}}}
\newcommand{\Dlate}{D_{\mathrm{late}}}
\newcommand{\CAP}{\mathrm{CAP}}
\newcommand{\srk}{\mathrm{srank}}

\newcommand{\finishdocument}{}

\titleformat{\section}{\Large\bfseries}{\thesection}{0.7em}{}
\titleformat{\subsection}{\large\bfseries}{\thesubsection}{0.7em}{}
\titleformat{\subsubsection}{\normalsize\bfseries}{\thesubsubsection}{0.7em}{}
\titlespacing*{\section}{0pt}{2.2ex plus 0.4ex}{0.8ex}
\titlespacing*{\subsection}{0pt}{1.8ex plus 0.3ex}{0.5ex}

\newcommand{\microsoftwordmark}{%
  \includegraphics[width=1.5in]{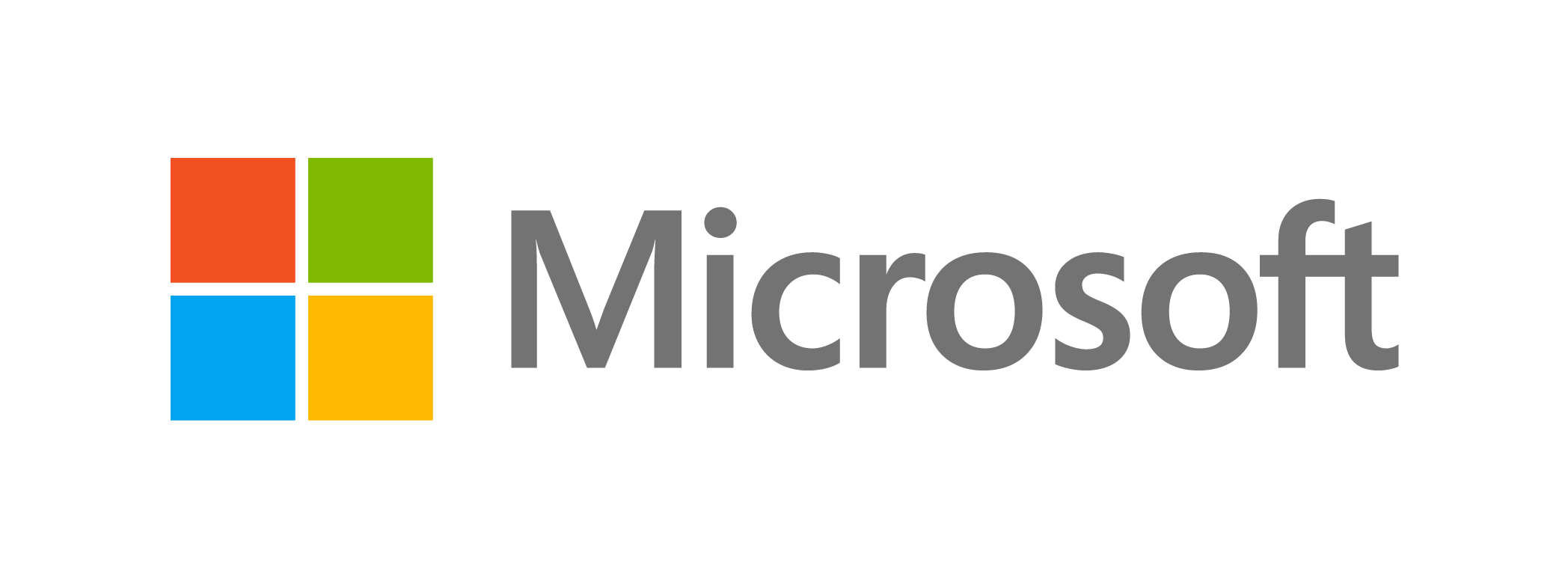}}
\newcommand{\positionedmicrosoftwordmark}{%
  \begin{tikzpicture}[overlay,baseline]
    \node[anchor=base west,inner sep=0pt,yshift=-44pt] at (0,0)
      {\microsoftwordmark};
  \end{tikzpicture}}

\fancypagestyle{preprintfirst}{%
  \fancyhf{}
  \fancyhead[C]{%
    \makebox[\headwidth][s]{%
      \positionedmicrosoftwordmark
      \hfill
      \raisebox{-30pt}[0pt][0pt]{2026 Aug}}}
  \fancyfoot[C]{\thepage}
  \renewcommand{\headrulewidth}{0.7pt}
  \renewcommand{\headrule}{\vskip 26pt\hbox to\headwidth{\color{black!65}\leaders\hrule height \headrulewidth\hfill}}
  \renewcommand{\footrulewidth}{0pt}
}
\fancypagestyle{preprint}{%
  \fancyhf{}
  \fancyfoot[C]{\thepage}
  \renewcommand{\headrulewidth}{0pt}
  \renewcommand{\footrulewidth}{0pt}
}
\makeatletter
\renewcommand{\maketitle}{%
  \thispagestyle{preprintfirst}
  \vspace*{8.35pt}
  {\centering
    {\fontsize{19}{23}\selectfont\bfseries \@title\par}
    \vspace{0.14in}
    {\large \@author\par}}
  \vspace{0.01in}
}
\makeatother

\renewenvironment{abstract}{%
  \begin{tcolorbox}[
    enhanced,
    colback=abstractbg,
    colframe=abstractbg,
    boxrule=0pt,
    arc=9pt,
    left=17pt,right=17pt,top=11pt,bottom=10pt]
}{%
  \vspace{6pt}
  {\color{black!20}\hrule height 0.5pt}
  \vspace{5pt}
  {\footnotesize
  Work done during an internship at Microsoft Research. Correspondence:
  Yifan Wang (\href{mailto:wang5617@purdue.edu}{wang5617@purdue.edu}), Xiaomin Li (\href{mailto:xiaominl@microsoft.com}{xiaominl@microsoft.com}).}
  \end{tcolorbox}
}

\begin{document}

\newif\ifcomments
 \commentsfalse     

\newcommand{\varun}[1]{%
  \ifcomments
    \textcolor{red}{[Varun: #1]}%
  \fi
}

\makeatletter
\newcommand{\arxivonly}{\ifdefined\ARXIVBUILD\expandafter\@firstofone\else\expandafter\@gobble\fi}
\newcommand{\iclronly}{\ifdefined\ARXIVBUILD\expandafter\@gobble\else\expandafter\@firstofone\fi}
\makeatother

\makeatletter
\newcommand{\markmainend}{\label{sec:mainend}\pdfsavepos\write\@auxout{\string\gdef\string\mainendpos{\the\pdflastypos}\string\typeout{MAINEND ypos=\the\pdflastypos sp page=\thepage}}}
\makeatother

\title{Model-Aware Data Selection from\\ In-and-Out Information Interplay}

\author{%
Yifan Wang$^{1}$ \quad Xiaomin Li$^{5}$ \quad Yuexing Hao$^{5}$ \quad Dongwon Jung$^{3}$ \quad Hemanth Neelgund Ramesh$^{2}$ \\ Ananth Grama$^{1}$ \quad Varun Chandrasekaran$^{4}$ \quad Yu Hu$^{5}$ \quad Andrzej Banburski-Fahey$^{5}$ \quad Jaron Lanier$^{5}$\\[0.12in]
{\color{black!58}
$^{1}$Purdue University \quad
$^{2}$University of Washington \quad
$^{3}$University of California, Davis \\
$^{4}$University of Illinois Urbana-Champaign\quad
$^{5}$Microsoft}%
}


\maketitle

\begin{abstract}
LLMs are effective representations that assimilate vast amounts of knowledge during pretraining, but post-training is necessary for models to reliably access this knowledge and “know what they know.” We observe an interesting rank equilibrium between knowledge stored in the weights and the data stream passing through the model. Across all model layers, we find that the hidden states (data stream) follow a U-shaped pattern, showing substantial compression in early layers and a steep rise during the late-layer decoding phase. In contrast, the weight rank follows an inverted U-shaped pattern, with very low rank in the early and late layers and high rank in the middle. We interpret this as an \emph{in-and-out information interplay}: intermediate activations do not need to carry content that the weights can supply later, so they primarily preserve what the weights cannot provide.
Motivated by this observation, we propose a model-aware data selection method, CAP (Counterfactual Assimilation Profile), which can determine whether a data candidate contains information accessible to the current model by utilizing the divergence gap in early- and late-layer representations between model-generated and reference responses. Across math, code, and science domains, CAP delivers \textbf{35.4\%} greater average improvement over the base model than the strongest baseline under different selection budgets. With only \textbf{10\%} of the data pool, CAP surpasses or matches full-pool training on math and science. We further show that CAP transfers to multimodal data selection and is robust to response horizon and noise.
\end{abstract}

\begin{figure}[H]
  \vspace{-0.2\baselineskip}
  \centering
  \setlength{\abovecaptionskip}{4pt}
  \setlength{\belowcaptionskip}{0pt}
  \begin{subfigure}[b]{0.63\linewidth}
    \centering
    \includegraphics[width=\linewidth]{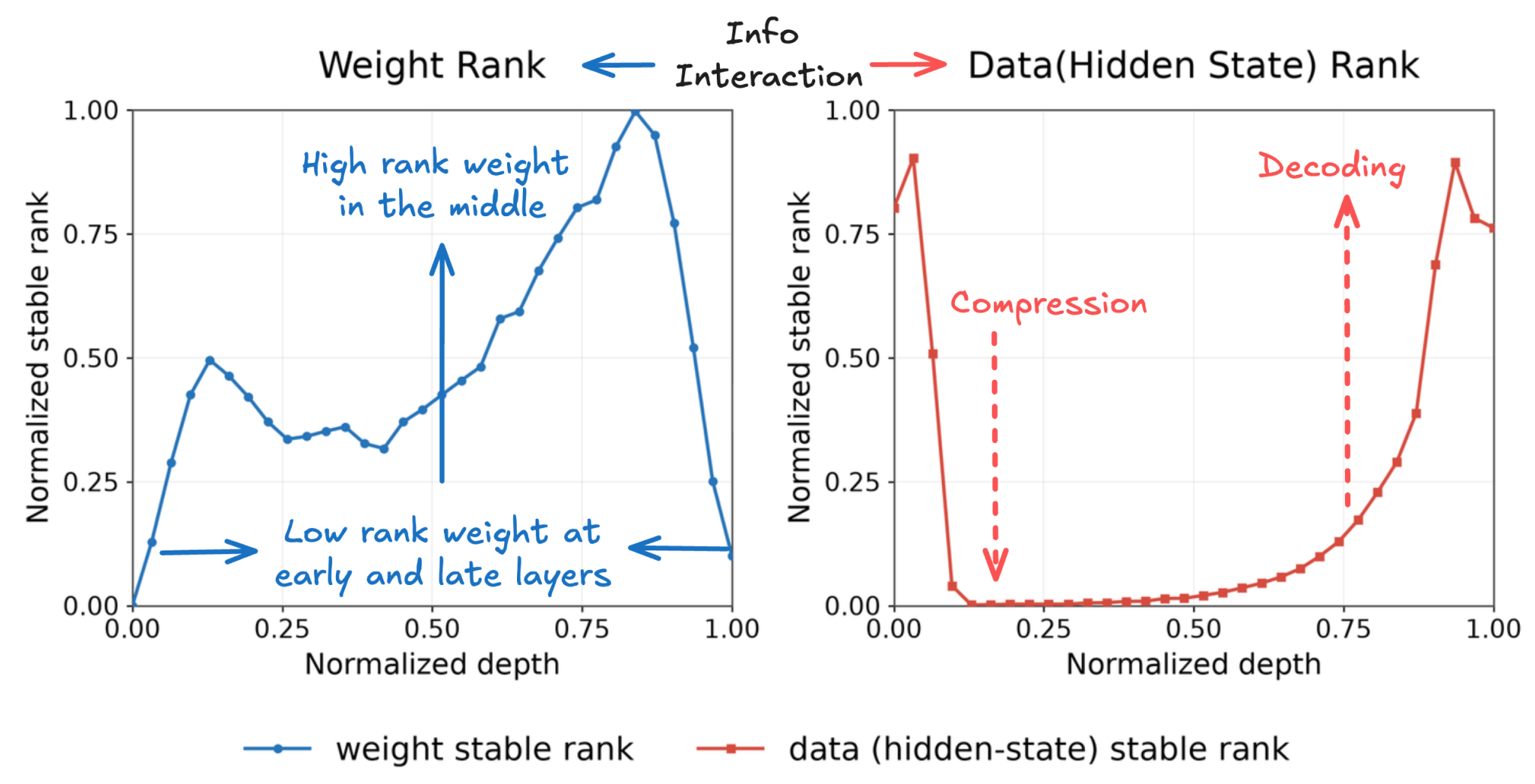}
    \caption{Rank equilibrium between weights and data.}
    \label{fig:ushape-annot}
  \end{subfigure}
  \hfill
  \begin{subfigure}[b]{0.35\linewidth}
    \centering
    \includegraphics[width=\linewidth]{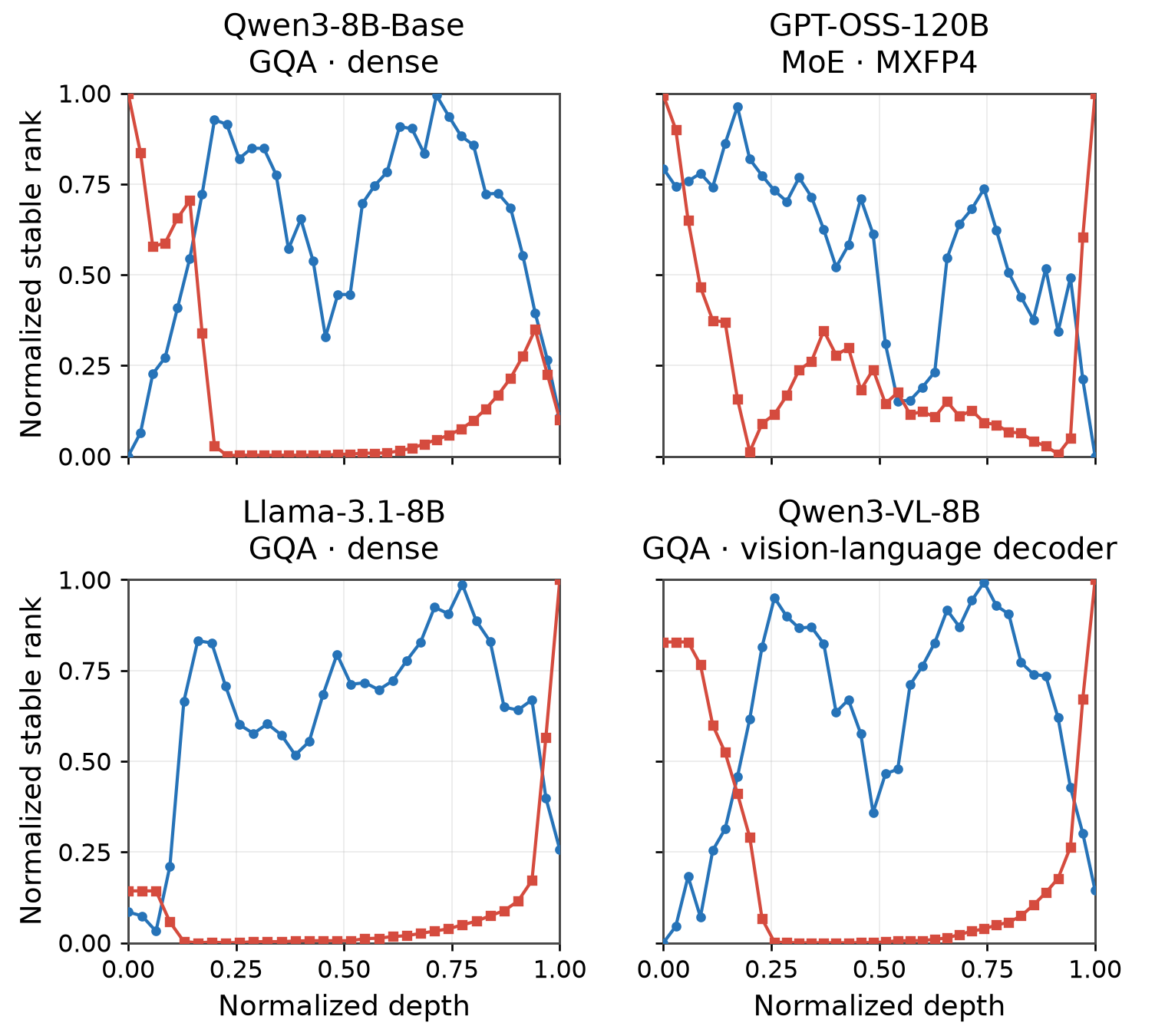}
    \caption{Cross-model consistency.}
    \label{fig:rank4}
  \end{subfigure}
  \caption{\textbf{The in-and-out information interplay.}
\textbf{(a)} Across model layers, weight stable rank (blue, circles) follows an inverted U-shape, while hidden-state stable rank (red, squares) follows the complementary U-shape: activations compress through the early and middle layers and re-expand near the output for decoding, as weight rank declines.
\textbf{(b)} This complementary pattern recurs across model families, scales, and architectures; the full 12-model survey is in Appendix~\ref{app:survey}.}
  \label{fig:ushape}
  \vspace{-0.3\baselineskip}
\end{figure}

\section{Introduction}
\label{sec:intro}

A modern language model acquires knowledge in two very different regimes. Pretraining assimilates a web-scale corpus into model weights \citep{deletang2024compression}: it determines what is \emph{stored}, but storage does not imply use, since a base model often fails at tasks whose ingredients are demonstrably represented in its parameters. Post-training changes much less about what is stored and much more about what is \emph{accessible}: which internal competencies can be surfaced, in what format, and under what instruction. Figure~\ref{fig:venn} makes this distinction explicit: the latent set is mainly expanded through (continued) pretraining, while post-training primarily expands the accessible set and can occasionally introduce new knowledge, with some knowledge also lost through forgetting \citep{kirkpatrick2017overcoming}.

Supervised fine-tuning (SFT) is the simplest and most widely used post-training stage, making the choice of SFT data a practical lever for improving accessibility. With strong open-weight base models, a few thousand curated examples can already produce competitive domain-specific assistants. This setting is particularly attractive in domains such as finance and medicine, where sensitive data often cannot leave the organization. Under limited compute and a large candidate pool of uneven utility, a small, well-chosen subset can outperform a much larger indiscriminate one \citep{zhou2023lima}. Which examples to keep therefore becomes a central question.

\newcommand{\figVenn}[4]{%
\begin{wrapfigure}{r}{#1\linewidth}
  \vspace{#2\baselineskip}
  \centering
  \includegraphics[width=\linewidth]{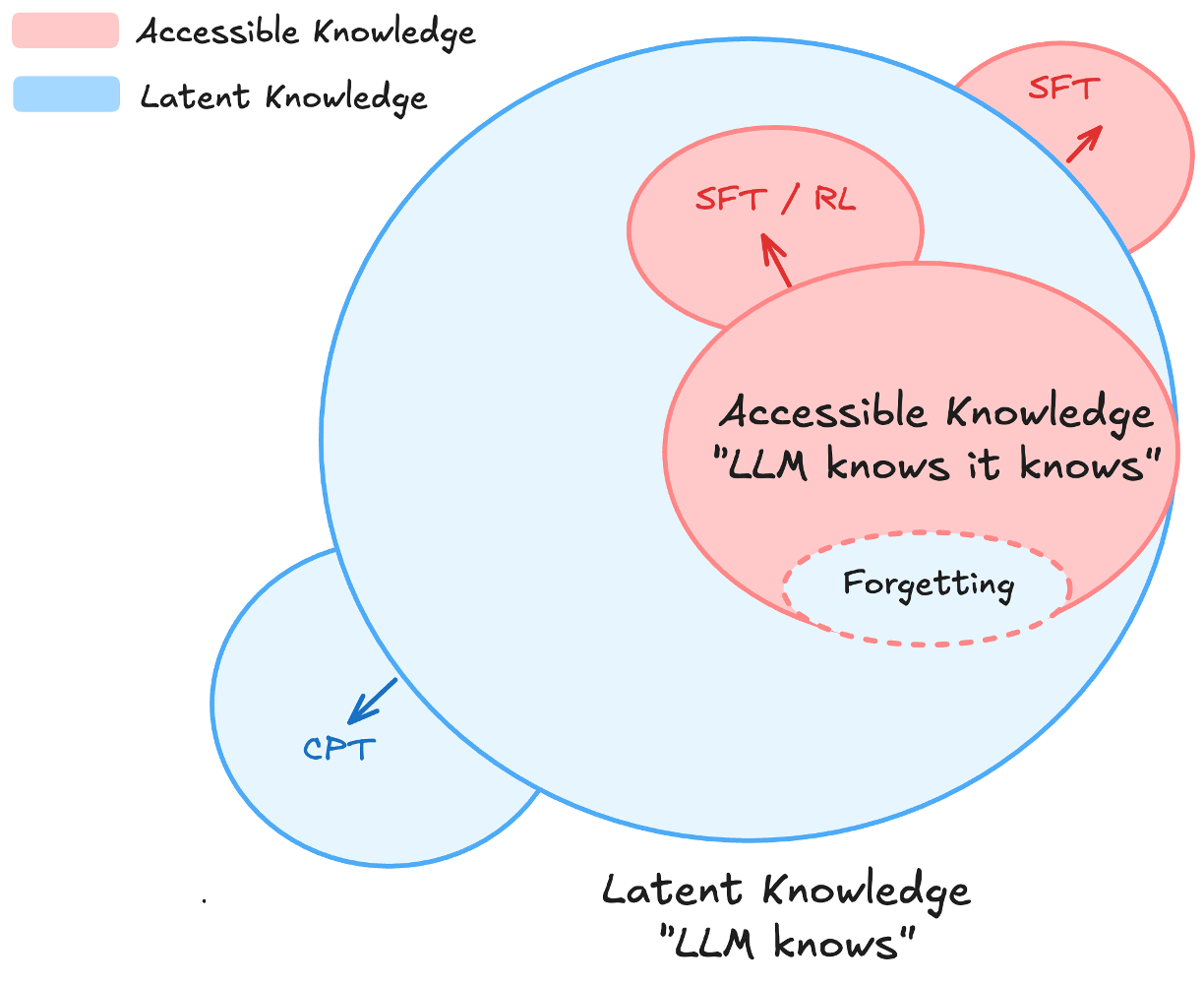}
  \vspace{#3\baselineskip}
  \caption{Latent and accessible knowledge.}
  \label{fig:venn}
  \vspace{#4\baselineskip}
\end{wrapfigure}}
\iclronly{\figVenn{0.45}{-1.2}{-1.3}{-1.2}}
\arxivonly{\figVenn{0.42}{-1.2}{-1.3}{-1.2}}

This decision is difficult because the value of an example depends on the model. An example the model can already reproduce provides little new signal, while one far outside its competence may provide mostly noise. Yet, most existing selection signals largely ignore the underlying model. Learned quality raters \citep{wettig2024qurating} score text in isolation, while diversity methods \citep{sener2018active} cover embedding space without asking what the model has already absorbed. Loss-based scores such as perplexity \citep{marion2023less} and IFD \citep{li2024ifd} consider the model, but reduce its response to a scalar that cannot distinguish \emph{knowledge I have not yet assimilated} from \emph{text that is intrinsically difficult, or simply junk}.

We identify this boundary of accessible knowledge inside the model through a pattern in how information is distributed between weights and activations. Measured layer by layer, trained decoders place high-rank weights in the middle of the stack, while hidden states compress through the middle and re-expand near the output (Figure~\ref{fig:ushape}). Neither profile is present at initialization; both emerge over pretraining and recur across model families, scales, attention mechanisms, dense and mixture-of-experts computation, and text-only and vision-language decoders. We interpret this as an \emph{in-and-out information interplay}: middle layers compress what comes \emph{in} against what is already stored, while late layers re-expand as stored knowledge is read \emph{out} for decoding. Deep layers are therefore where information brought by the data meets knowledge already stored in the weights.

This observation motivates our approach, \emph{Counterfactual Assimilation Profile} (CAP). For a candidate $(x,y)$, we run the model on the real pair $[x;y]$ and on the counterfactual pair $[x;\hat y]$, where $\hat y$ is the model's own generation, and measure the per-layer divergence between the two runs. Shallow-layer divergence captures surface mismatch such as format and style, while deep-layer divergence captures mismatch at locations where stored knowledge is retrieved. CAP residualizes deep divergence on shallow divergence, isolating information that the model has not yet assimilated, precisely the information SFT should make accessible. Structureless noise contains no retrievable content, so it provably falls to the bottom of the CAP ranking rather than the top, where loss-based scores may place it. Building on these insights, we make the following contributions:

\begin{enumerate}
  \item \textbf{Phenomenon and theory.} We identify a coupled rank equilibrium between weights and hidden states that emerges over pretraining and holds across architectures, scales, and modalities. In a memory-augmented computation model, we show that the observed U-shaped and inverted-U-shaped profiles arise as optimal capacity profiles, obey a per-depth conservation law, and make the trough of the activation U-shape a measure of assimilated knowledge (Section~\ref{sec:interplay}).

\item \textbf{Performance and data efficiency.}
Across math, code, and science and across selection budgets on Marin-8B-Base, CAP delivers 35.4\% greater average improvement over the base model than the strongest baseline, and nearly $1.9\times$ the gain of random selection. (Sections~\ref{sec:cap} and \ref{sec:analysis}).

  \item \textbf{Generality and robustness.} CAP transfers unchanged to a vision-language decoder and to finance and medicine, without requiring modality-specific scoring machinery. It remains effective across nearly an order of magnitude of response horizons and rejects injected noise that perplexity ranks above the entire clean distribution (Sections~\ref{sec:multimodal} and \ref{sec:analysis}).
\end{enumerate}
\section{Related Work}
\label{sec:related}

\paragraph{Data selection for post-training.}
Work on choosing post-training data splits into roughly three directions, distinguished by where the signal comes from. \emph{External scoring} judges an example without considering the current model: learned quality raters such as QuRating \citep{wettig2024qurating}, LLM-as-judge pipelines such as AlpaGasus \citep{chen2024alpagasus}, and complexity/quality composites such as DEITA \citep{liu2024deita}. \emph{Geometric and distributional} methods select for coverage or for match to a target distribution: k-center coreset selection \citep{sener2018active}, DSIR-style importance resampling \citep{xie2023dsir}, and deduplication or clustering pipelines. \emph{Model-internal} methods read the current target model: perplexity and loss-based pruning \citep{marion2023less}, instruction-following difficulty \citep{li2024ifd}, and gradient-influence matching such as LESS \citep{xia2024less}. CAP belongs to the third family but differs from others in what it reads. Loss-based scores compress the model's state into a single likelihood number, and gradient methods require a target eval set and backward passes. CAP instead contrasts two forward passes, the reference response against the model's own continuation, and localizes the contrast by depth, which is what lets it separate unassimilated content from surface difficulty. That separation is also what our theory predicts, and what makes noise rejection provable rather than incidental.

\paragraph{Compression, rank, and structure across depth in LLMs and VLMs.}
Our observation sits in a line of work on the low-dimensional and highly structured nature of transformer computation. Language modeling as compression \citep{deletang2024compression} provides the framing; fine-tuning within a low intrinsic dimension \citep{aghajanyan2021intrinsic} and low-rank adaptation \citep{hu2022lora} show that task-relevant updates occupy few directions; the low-rank simplicity bias of gradient descent \citep{huh2023lowrank} explains why; and post-training compression methods such as SparseGPT \citep{frantar2023sparsegpt} and Wanda \citep{sun2024wanda} exploit the resulting redundancy, empirically showing that layers differ sharply in how much compression they tolerate. Complementary phenomenological work shows that activations are far from generic: attention sinks concentrate mass on a few positions \citep{xiao2024streamingllm} and massive activations dominate a handful of hidden dimensions \citep{sun2024massive}, both varying systematically with depth. Interpretability work in the same vein reads intermediate layers out directly, as in logit and tuned lenses \citep{belrose2023tunedlens}, and localizes factual recall to middle-layer MLPs \citep{meng2022rome, dai2022knowledge}, consistent with our finding that the interior is where stored knowledge is matched against the input. The same depth-wise structure appears in vision-language decoders \citep{deitke2024molmo}. What is new here is the \emph{coupling}: weight rank and hidden-state rank form complementary profiles, the coupling strengthens over pretraining, and it can be converted into a practical, model-aware data selection signal.

\section{In-and-Out Information Interplay: Rank Equilibrium Between Weights and Data}
\label{sec:interplay}

\subsection{The Observation}
\label{sec:obs}

We measure the stable rank
$\srk(A) = \|A\|_F^2 / \|A\|_2^2$
of weight matrices and token hidden-state matrices at each layer, normalized by dimension. Three facts emerge: (i) \emph{The profiles are coupled:} middle layers hold high-rank weights and low-rank activations, while late layers reverse the pattern (Figure~\ref{fig:ushape-annot}). (ii) \emph{The coupling is learned:} we observe that the complementary profiles emerge progressively over pretraining rather than being present at initialization. (iii) \emph{The coupling is not model-specific:} the same qualitative structure appears across all models we examine. Figure~\ref{fig:rank4} spans dense, mixture-of-experts, and vision-language decoders, while the full survey covers 12 models from 3.8B to 120B parameters (Figure~\ref{fig:rank-survey}). Activation compression and late re-expansion are the most consistent features; weight profiles are noisier but generally concentrate rank in the interior.

We interpret this as an interplay between what flows \emph{in} through the activations and what is stored in the weights. Middle layers compress activations while stored information can be supplied later, and late layers re-expand as that information is retrieved for decoding. This suggests a model-aware signal for data selection: assimilated content should induce little deep-layer change relative to the model's own generation, while unassimilated content should induce substantially more. Shallow layers should primarily reflect surface mismatch. This motivates a counterfactual comparison, a depth split, and residualization of deep divergence on shallow divergence.

\subsection{Theoretical Analysis}
\label{sec:theory}

We formalize this equilibrium with a simple capacity model of computation against memory. The full model, assumptions, mapping to transformer measurements, and proofs are given in Appendix~\ref{app:theory}.

A depth-$L$ computation carries activation capacity $r_\ell$ and spends weight capacity $w_\ell$ at each layer. A memory holds $K$ units of task-relevant knowledge, and a retrieval schedule $k_\ell$ determines when that knowledge is injected. Activations must carry content that will not be supplied later, together with an address for content that will be retrieved later. We assume that addressing stored content is cheaper than carrying it directly. The training compression bias is represented by minimizing $\sum_\ell r_\ell$, with total weight capacity as a tiebreaker.

\begin{theorem}[Shape of the optimal profiles]
\label{thm:shape}
Under Assumptions~\ref{ass:addr}--\ref{ass:match} (Appendix~\ref{app:model}), the optimal computation exists, is unique, and satisfies, up to one fractional boundary layer:
\begin{itemize}
  \setlength{\itemsep}{1pt}
  \item[(i)] \textbf{Deferred retrieval:} $K_{>\ell} = \min\{K,\,(L-\ell)B_r\}$, so retrieval is pushed as late as the bandwidth $B_r$ permits.
  \item[(ii)] \textbf{U-shaped activations:} $r_\ell = \max\{H - B_t\ell,\ g(K_{>\ell})\}$ descends at the maximal rate $B_t$ from $H$ to the trough $r^\star = g(K) = c(R-K) + \varphi(K)$, holds there, and rises back to $cR$ at the output.
  \item[(iii)] \textbf{Interior-concentrated matching:} the matching component of $w_\ell$ vanishes at both boundaries and attains its maximum exactly under the activation trough, at a value strictly increasing in $K$; in the matching-dominated regime of Assumption~\ref{ass:regime}, the full weight profile $w_\ell$ peaks in the interior.
\end{itemize}
\end{theorem}

Parts (i) and (ii) arise because the computation compresses as quickly as permitted and delays retrieval until needed. Part (iii) additionally requires matching cost to grow with both the amount of stored content and the degree of compression, yielding an interior-peaked weight profile in the matching-dominated regime.

\begin{theorem}[In-and-out conservation law]
\label{thm:conservation}
For every feasible computation and every depth $\ell$, $r_\ell + c\,K_{>\ell} \ge cR$, and consequently $r_\ell + \frac{c}{\sigma}\sum_{j>\ell} w_j \ge cR$. On the retrieval arm of the optimal computation,
\[
r_\ell + c\,K_{>\ell} = cR + \varphi(K_{>\ell}).
\]
\end{theorem}

Thus, at each depth, information carried by the activations plus information that can still be supplied from memory must meet the task requirement.

\begin{corollary}[The trough is a knowledge meter]
\label{cor:emergence}
Let $\psi(K) = cK - \varphi(K)$. Then $\psi(0)=0$, $\psi' \ge \eta > 0$, and
\[
cR - r^\star(K) = \psi(K).
\]
Hence the trough deepens monotonically as retrievable knowledge grows, and at $K = 0$ the optimal profile is non-increasing in depth.
\end{corollary}

This is consistent with our observation that the coupled profiles emerge over pretraining and predicts that domain-specific continued pretraining should deepen the trough on in-domain data.

\paragraph{The assimilation gap.}
\label{sec:theory_gap}
For a candidate $(x,y)$, let $K_y$ denote the storable content of the reference response and $K_y^{\cap} \le K_y$ the part already retrievable from the weights given $x$. We define
\begin{equation}
\label{eq:gap}
  G(x,y) \;:=\; \psi(K_y) - \psi(K_y^{\cap}) \;\ge\; 0 .
\end{equation}
The gap is zero both when the model already retrieves the relevant content of $y$ and when $y$ contains no storable structure. Thus $G$ captures unassimilated but learnable content, which scalar loss alone cannot isolate.

\paragraph{From the gap to a measurable score.}
Run the model on $[x;y]$ and on the counterfactual $[x;\hat y]$, and let $D_\ell$ denote their divergence at depth $\ell$ (Section~\ref{sec:method}). The deferred-retrieval structure motivates an exclusion restriction: shallow divergence should primarily reflect surface mismatch, while deep divergence can additionally reflect the assimilation gap. We model
\[
\Dearly = aS + \varepsilon_e,
\qquad
\Dlate = a'S + bG + \varepsilon_l,
\]
where $S$ is surface mismatch between the reference and the model's own generation. Assumption~\ref{ass:factor} defines the factor model and the resulting surface-orthogonal gap $G_\perp$.

\begin{theorem}[Identification of the surface-orthogonal gap]
\label{thm:ident}
Under Assumption~\ref{ass:factor}, the population regression residual
\[
\CAP = \Dlate - \alpha - \beta\,\Dearly
\]
has slope
\[
\beta = (1-\delta)(a' + b\gamma)/a
\]
and satisfies
\[
\CAP
=
b\,G_\perp
+
\delta\,(a' + b\gamma)\,(S - \mathbb{E}S)
-
\beta\,\varepsilon_e
+
\varepsilon_l .
\]
Consequently: (i) $\mathrm{Cov}(\CAP, G_\perp) = b\,\mathrm{Var}(G_\perp) > 0$; (ii) if $\sigma_{\varepsilon_e} = 0$, then $\CAP = b\,G_\perp + \varepsilon_l$; and (iii) surface difficulty leaks into CAP only through the attenuation term $\delta$.
\end{theorem}

CAP therefore ranks by unassimilated content after removing the component predictable from surface difficulty.

\begin{corollary}[Noise rejection]
\label{cor:noise}
Let a contaminating example have $G = 0$ and surface value $s$, and suppose that it follows the same structural equations as the clean corpus. Then
\[
\mathbb{E}\left[\CAP \,\middle|\, S = s,\ G = 0\right]
  =
  \big[\,\delta\,a' - (1-\delta)\,b\gamma\,\big]\,(s - \mathbb{E}S)
  - b\,\mathbb{E}G .
\]
If $\gamma > 0$ and $\delta < b\gamma/(a' + b\gamma)$, this expectation decreases with $s$, so high-surface-difficulty noise receives a low CAP score.
\end{corollary}

This makes noise rejection a consequence of residualization rather than a particular distance metric. Proposition~\ref{prop:select} further shows that top-$m$ CAP selection retains sufficiently high-gap examples with failure probability exponentially small in the gap margin.

The theory characterizes the qualitative structure of the observed profiles and their connection to data selection. The activation U-shape and deferred retrieval schedule depend on the addressing and bandwidth assumptions, while the interior-peaked weight profile additionally requires the matching assumptions. Stable rank is a measurable proxy for capacity, and Theorem~\ref{thm:ident} is conditional on the factor model. The theory does not determine the empirical depth split, response horizon $R$, or downstream effect size; Appendix~\ref{app:scope} discusses these points in detail.
\section{CAP: Counterfactual Assimilation Profile for Model-Aware Data Selection}
\label{sec:cap}

\subsection{Method}
\label{sec:method}

\paragraph{Inputs.}
For a candidate with prompt $x$ and reference response $y$, we form the real sequence $[\,x;y\,]$ and the counterfactual $[\,x;\hat y\,]$, where $\hat y$ is the model's own greedy generation for the same prompt. Both responses are truncated to $R$ tokens ($R=48$ by default; ablated in Section~\ref{sec:ablate_R}). The generated response reflects what the model can produce from its current knowledge, so differences between the two runs isolate information introduced by the reference response.

\paragraph{Divergence and depth split.}
Let $H_\ell^{\mathrm{real}}$ and $H_\ell^{\mathrm{gen}}$ denote the layer-$\ell$ hidden states at aligned response positions in an $L$-layer model. Guided by the deferred-retrieval structure of Theorem~\ref{thm:shape}, we summarize the per-layer divergence with shallow and deep averages:
\begin{equation}
  D_\ell = 1 - \cos\!\big(H_\ell^{\mathrm{real}},\, H_\ell^{\mathrm{gen}}\big),
  \qquad
  \Dearly = \operatorname{mean}_{\ell \le 0.33L} D_\ell,
  \qquad
  \Dlate = \operatorname{mean}_{\ell \ge 0.66L} D_\ell ,
\end{equation}
with $D_\ell$ averaged over the $R$ aligned response positions. Shallow divergence primarily captures surface mismatch, while deep divergence additionally reflects differences where stored knowledge is retrieved.

\paragraph{Score and selection.}
Surface difficulty can inflate both quantities, so we residualize:
\begin{equation}
  \CAP = \Dlate - (\alpha + \beta\,\Dearly),
\end{equation}
where $\alpha$ and $\beta$ are fit by ordinary least squares of $\Dlate$ on $\Dearly$ over the candidate pool. By Theorem~\ref{thm:ident}, CAP captures the component of deep divergence not explained by surface difficulty, corresponding to the surface-orthogonal assimilation gap up to scale and noise. We z-score CAP within each domain and select the highest-scoring examples under the budget. Scoring requires one greedy generation and two forward passes per candidate, with no auxiliary rater, backward pass, or target set.

\subsection{Experimental Setup}
\label{sec:setup}

We use Marin-8B-Base \citep{marin2025} for all key experiments. Marin is developed fully in the open: its training code, checkpoints, training process, and the sources of its complete pretraining corpus are publicly released. This lets us construct SFT candidate pools from data sources outside its pretraining mixture, substantially reducing pretraining contamination as a confounder in evaluating data selection.

We fine-tune Marin-8B-Base with full-parameter SFT for one epoch using AdamW, a peak learning rate of $1\times 10^{-5}$, a cosine schedule, and a 2048-token context. Every method shares the same candidate pool, budget, data order, seed, optimizer, and training configuration, so only the selected subset differs (Appendix~\ref{app:training}).

Each domain uses a deduplicated pool of 80k examples. \emph{Math} selects from NuminaMath-CoT \citep{numina2024} and evaluates GSM8K strict-match accuracy \citep{cobbe2021gsm8k}; \emph{code} selects from OpenCoder opc-sft-stage2 \citep{huang2024opencoder} and evaluates HumanEval pass@1 \citep{chen2021codex}; and \emph{science} selects from the StackExchange split of WebInstructSub \citep{yue2024mammoth2} and evaluates MMLU-Pro STEM with 5-shot CoT \citep{wang2024mmlupro}. Baselines span the major signal families in Section~\ref{sec:related}: random selection, perplexity (PPL; \citealp{marion2023less}), IFD \citep{li2024ifd}, QuRating \citep{wettig2024qurating}, gradient-based LESS \citep{xia2024less}, and k-center \citep{sener2018active}.

\begin{table}[tb]
  \centering
  \small
  \setlength{\tabcolsep}{5pt}
  \begin{tabular}{lll}
    \toprule
    Method & Signal family & Signal \\
    \midrule
    Random & Heuristic & none \\
    PPL \citep{marion2023less} & Model-internal (loss) & absolute model loss \\
    IFD \citep{li2024ifd} & Model-internal (difficulty) & cond.\ vs.\ uncond.\ loss ratio \\
    QuRating \citep{wettig2024qurating} & External scoring & learned quality rater \\
    LESS \citep{xia2024less} & Model-internal (gradient) & target-set gradient influence \\
    k-center \citep{sener2018active} & Geometric coverage & embedding geometry \\
    \midrule
    CAP (ours) & Model-internal (repr.) & depth-split counterfactual divergence \\
    \bottomrule
  \end{tabular}
  \caption{Data-selection baselines spanning heuristic, model-internal, external-scoring, and geometric signals.}
  \label{tab:baselines}
\end{table}

\subsection{Main Results}
\label{sec:main_results}

\begin{table}[H]
  \centering
  \small
  \setlength{\tabcolsep}{4pt}
  \resizebox{\textwidth}{!}{%
  \begin{tabular}{lcccc c cccc c cccc c c}
    \toprule
    & \multicolumn{4}{c}{\textbf{Math} (GSM8K)} &
    & \multicolumn{4}{c}{\textbf{Code} (HumanEval)} &
    & \multicolumn{4}{c}{\textbf{Science} (MMLU-Pro-5)} &
    & \textbf{Avg. @8k} \\
    \cmidrule(lr){2-5}
    \cmidrule(lr){7-10}
    \cmidrule(lr){12-15}
    Method
    & 1k & 2k & 4k & 8k
    & & 1k & 2k & 4k & 8k
    & & 1k & 2k & 4k & 8k
    & & \\
    \midrule

    Base model
    & \multicolumn{4}{c}{61.11}
    & & \multicolumn{4}{c}{30.49}
    & & \multicolumn{4}{c}{32.55}
    & & 41.38 \\

    \midrule

    Random
    & \textbf{63.99} & \underline{65.88} & 68.01 & 69.37
    & & 28.05 & \textbf{32.93} & \underline{32.32} & 30.49
    & & 33.89 & 35.50 & \underline{37.57} & 38.19
    & & 46.02 (+4.63) \\

    PPL
    & 62.02 & 64.97 & 65.88 & 68.92
    & & \underline{30.49} & 28.05 & 28.05 & \underline{32.93}
    & & \textbf{34.83} & 34.87 & 36.41 & 37.72
    & & 46.52 (+5.14) \\

    IFD
    & 61.41 & 62.70 & 65.05 & 66.34
    & & 25.00 & 28.66 & 26.83 & 28.66
    & & 33.85 & 34.56 & 37.48 & 38.27
    & & 44.42 (+3.04) \\

    QuRating
    & 61.11 & 63.76 & 67.78 & 66.11
    & & 29.27 & 28.66 & 31.71 & \underline{32.93}
    & & 34.14 & 35.06 & \textbf{38.76} & \underline{38.51}
    & & 45.85 (+4.47) \\

    LESS
    & 60.27 & 64.29 & \underline{68.16} & \underline{70.81}
    & & \textbf{31.71} & \underline{32.32} & \textbf{33.54} & \textbf{35.37}
    & & 34.57 & \underline{36.03} & 36.59 & 38.39
    & & \underline{48.19} (+6.81) \\

    k-center
    & \underline{63.31} & 64.44 & 66.11 & 70.58
    & & \underline{30.49} & 27.44 & 29.27 & 32.32
    & & 33.78 & 35.07 & 36.29 & 37.08
    & & 46.66 (+5.28) \\

    \midrule

    CAP (ours)
    & 62.32 & \textbf{67.78} & \textbf{74.30} & \textbf{76.04}
    & & \textbf{31.71} & 29.88 & \textbf{33.54} & \textbf{35.37}
    & & \underline{34.66} & \textbf{36.23} & 37.26 & \textbf{39.05}
    & & \textbf{50.15 (+8.77)} \\

    \bottomrule
  \end{tabular}}
  \caption{Accuracy after SFT on subsets of the 80k-example pools. The final column reports the macro-average accuracy at the 8k budget, with improvement over the base model shown in parentheses. \textbf{Best} and \underline{second best} are marked per column. Results with standard deviations are in Appendix~\ref{app:full_results}.}
  \label{tab:main}
\end{table}

\begin{figure}[H]
  \centering
  \includegraphics[width=0.32\linewidth]{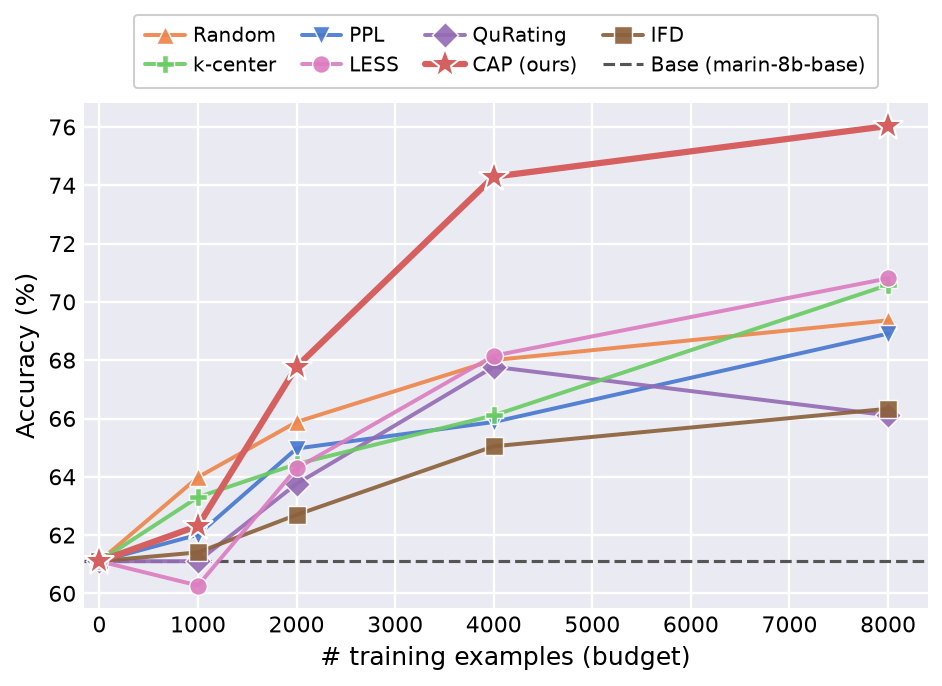}\hfill
  \includegraphics[width=0.32\linewidth]{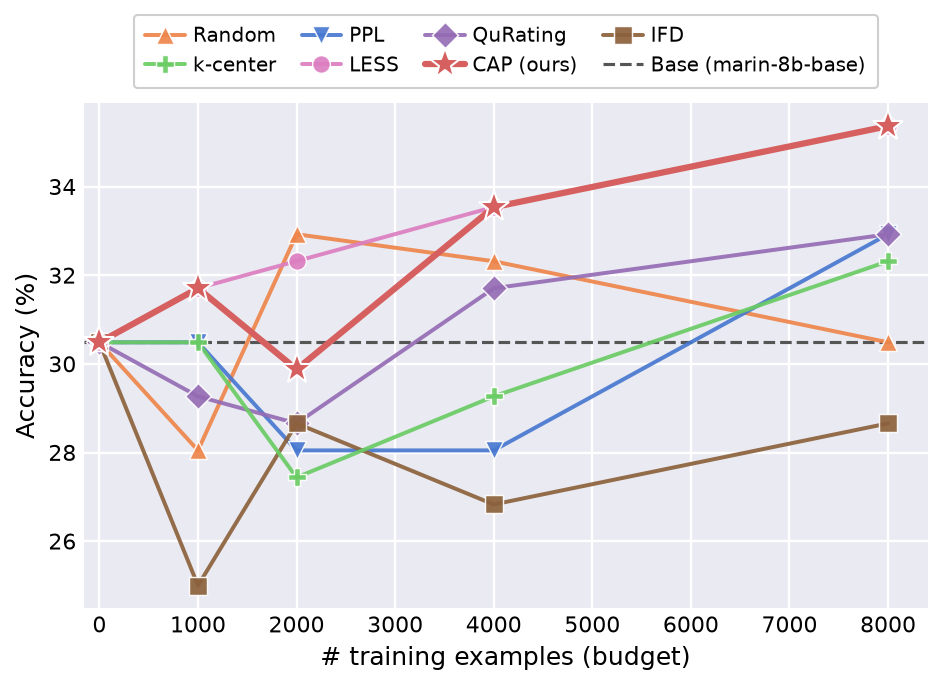}\hfill
  \includegraphics[width=0.32\linewidth]{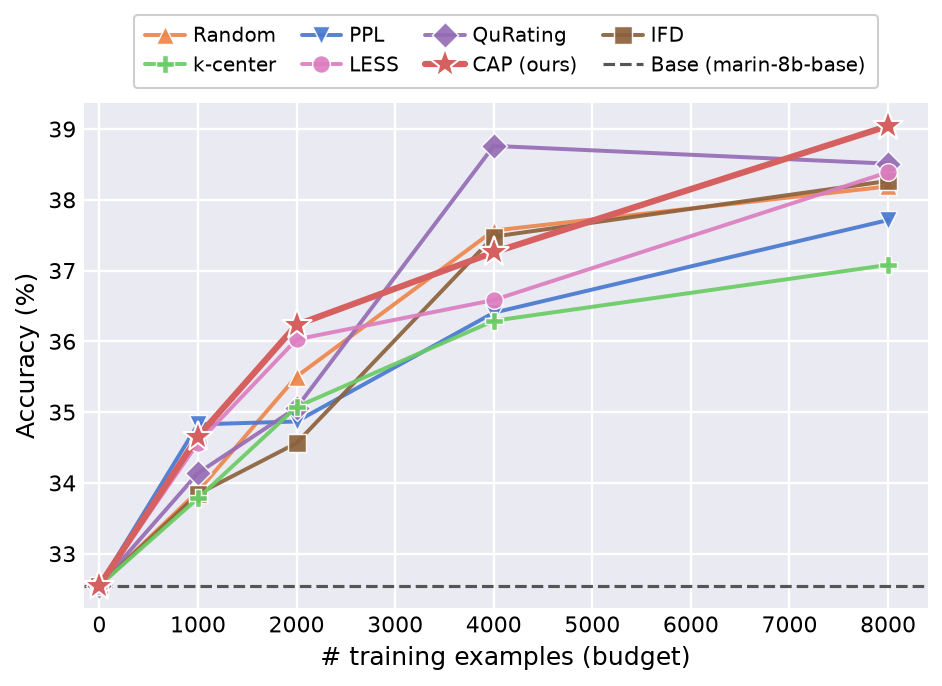}
  \caption{Mean accuracy versus selection budget on math (left), code (middle), and science (right). Shaded regions show one standard deviation.}
  \label{fig:budget_curves}
\end{figure}

Across math, code, and science and across all four selection budgets, CAP provides the strongest overall performance. Averaged over all 12 domain-budget settings, its improvement over the base model is 35.4\% larger than that of LESS, the strongest single baseline on average. CAP wins or ties for first in 8 of the 12 settings (Table~\ref{tab:main}, Figure~\ref{fig:budget_curves}).

At the 8k budget, CAP gains an average of $+8.77$ points over the base model, compared with $+6.81$ for LESS and $+4.63$ for random selection. Thus, at this budget, CAP achieves nearly $1.9\times$ the gain of random selection. On math, CAP reaches $+14.93$ points over the base model, compared with $+9.70$ for the strongest baseline, with the margin growing to more than five points at 4k and 8k. On code, CAP ties LESS at 8k, while IFD ends below the base model, illustrating that difficulty alone is not a reliable measure of training value. Science is the most competitive domain: CAP leads at 2k and 8k, while PPL and QuRating lead at 1k and 4k, respectively.

\paragraph{Data efficiency relative to full-pool training.}
With only 8k examples, or 10\% of the candidate pool, CAP reaches 76.04 on GSM8K, exceeding full-pool training at 72.48. On science, the same 10\% subset reaches 39.05, within one point of full-pool training at 39.82. The 8k CAP models also outperform Marin-8B-Instruct on both math and science, despite the instruct model being trained on a substantially broader mixture. Code is the exception: full-pool training reaches 43.29, indicating that this domain benefits more from data volume. Even there, CAP maintains a consistent advantage over random and PPL as the budget increases to 60k examples (Appendix~\ref{app:code_scale}).

\subsection{Generalizability: Multimodal and Further Domains}
\label{sec:multimodal}

Because CAP uses only hidden states and the model's own generation, it does not assume text-only inputs. Several baselines do: QuRating relies on a text-trained rater, IFD's unconditioned loss is not directly defined when the instruction contains an image, and k-center requires an embedding space that represents the visual component.

\begin{wraptable}{r}{0.48\linewidth}
  \vspace{-0.8\baselineskip}
  \centering
  \small
  \setlength{\tabcolsep}{4pt}
  \begin{tabular}{lcccc}
    \toprule
    Method & 1k & 2k & 4k & 8k \\
    \midrule
    Base (no SFT) & \multicolumn{4}{c}{0.00} \\
    \midrule
    Random & \textbf{30.77} & 29.12 & \underline{35.71} & 40.66 \\
    PPL & 26.92 & \underline{32.97} & \underline{35.71} & \underline{41.76} \\
    CAP (ours) & 26.92 & \textbf{34.62} & \textbf{40.66} & \textbf{46.15} \\
    \bottomrule
  \end{tabular}
  \caption{Multimodal selection on Molmo2-O-7B: held-out document-split accuracy from a 20k candidate pool.}
  \label{tab:multimodal}
  \vspace{-0.8\baselineskip}
\end{wraptable}

We apply CAP unchanged, including $R=48$, to the \emph{pretrain} checkpoint of Molmo2-O-7B \citep{clark2026molmo2}, ensuring that the target domain is not already part of its post-training mixture. The candidate pool contains 20k deduplicated document-understanding examples from Molmo2-SynMultiImageQA (Appendix~\ref{app:training}). CAP trails random at 1k, where all methods operate in a high-variance regime, but leads from 2k onward. At 8k it reaches 46.15, outperforming the runner-up by 4.39 points and random selection by 5.49 points (Table~\ref{tab:multimodal}).

\paragraph{Further specialized domains (finance, medicine).}

We further evaluate CAP on finance and medicine to test whether its model-aware signal remains effective in more specialized domains. CAP transfers unchanged to both settings and leads on every reported metric, achieving a 13.8\% relative improvement over PPL on finance and an 8.2\% improvement on medicine. The gains span both domain knowledge and reasoning-oriented evaluations (Appendix~\ref{app:domains}).

\section{Analysis and Ablation}
\label{sec:analysis}

\subsection{Selection Cost versus Performance}
\label{sec:cost}

\begin{figure}[ht]
  \centering
  \includegraphics[width=0.90\linewidth]{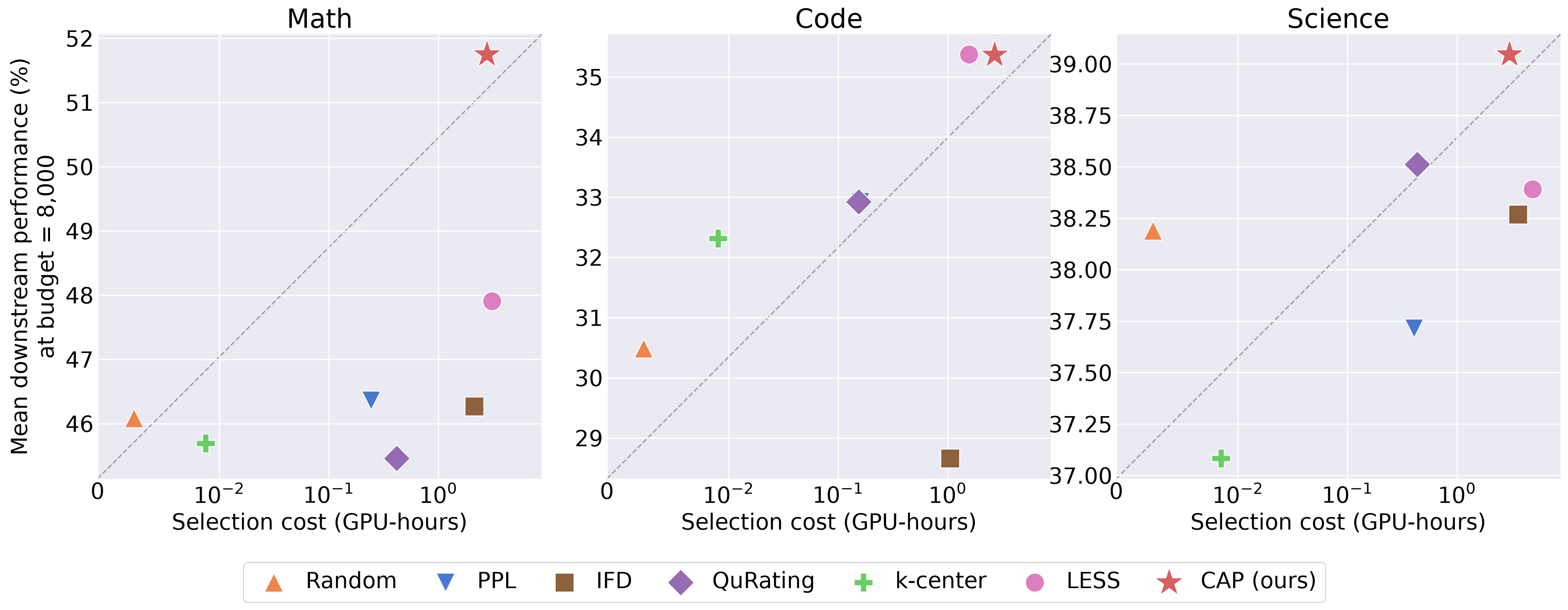}
  \caption{Accuracy at the 8k budget versus selection cost in utilization-weighted GPU-hours. The symmetric log scale retains the zero-cost random baseline.}
  \label{fig:cost}
\end{figure}

CAP requires one greedy generation and two forward passes per candidate, with no auxiliary rater training, backward pass, or target set. Figure~\ref{fig:cost} places the methods on the cost--performance plane at the 8k budget, with cost measured in utilization-weighted GPU-hours to discount allocated time during which GPUs are idle (Appendix~\ref{app:cost}). Averaged across math, code, and science, CAP uses $2.814$ GPU-hours, compared with $3.175$ for LESS and $2.259$ for IFD. CAP is therefore $11.4\%$ cheaper than LESS but $24.6\%$ more expensive than IFD, and is also more expensive than PPL, QuRating, and k-center. This additional scoring cost is offset by stronger data efficiency: random selection requires 8k examples to reach 69.37 on GSM8K, whereas CAP reaches the same level at roughly 2.5k examples by interpolation between its 2k and 4k results, corresponding to a $3.2\times$ reduction in selected data.

\arxivonly{\needspace{8\baselineskip}}
\subsection{Ablation: The Response-Length Parameter \texorpdfstring{$R$}{R}}
\label{sec:ablate_R}

\begin{wraptable}{r}{0.43\linewidth}
  \vspace{-0.8\baselineskip}
  \centering
  \small
  \setlength{\tabcolsep}{4pt}
  \begin{tabular}{lccc}
    \toprule
    $R$ & Math & Code & Science \\
    \midrule
    16  & 71.95 & 32.32 & 38.88 \\
    48  & \textbf{76.04} & \textbf{35.37} & 39.05 \\
    144 & 72.93 & 31.71 & \textbf{39.72} \\
    \midrule
    Random & 69.37 & 30.49 & 38.19 \\
    Best baseline & 70.81 & 35.37 & 38.51 \\
    \bottomrule
  \end{tabular}
  \caption{Effect of response horizon $R$ at the 8k budget.}
  \label{tab:ablate_R}
  \vspace{-0.8\baselineskip}
\end{wraptable}

Beyond the depth split, CAP has one numerical hyperparameter: the response horizon $R$ over which the real and counterfactual runs are compared. Table~\ref{tab:ablate_R} reruns the full pipeline at three horizons spanning nearly an order of magnitude. Every horizon outperforms random selection in all three domains, while CAP exceeds the strongest baseline in six of the nine horizon-domain settings and matches it in one more. The default $R=48$ performs best on math and code, exceeding the strongest math baseline by $5.23$ points and matching LESS on code. Science instead continues to improve with a longer horizon, reaching $39.72$ at $R=144$. Short horizons may miss reasoning expressed later in the response, while long horizons can include divergence caused by accumulated generation differences rather than assimilation. We therefore use $R=48$ as the balanced shared setting across domains.

\subsection{Noise Rejection}
\label{sec:noise}

High loss can arise for two different reasons: an example may contain useful knowledge the model has not yet assimilated, or it may simply be difficult because it lacks learnable structure. Difficulty-based scores cannot distinguish these cases, since both can produce high loss. To test whether CAP separates them, we inject 5\% random junk into a clean pool and score the resulting mixture with Marin-8B-Base. Perplexity ranks the junk above the entire clean distribution, whereas CAP ranks it below 99.7\% of clean examples (Figure~\ref{fig:noise}). This behavior is consistent with Corollary~\ref{cor:noise}: structureless noise has no assimilation gap, so although it appears difficult at the surface level, its deep divergence falls below what the corpus regression predicts from that difficulty. CAP therefore suppresses such contamination without requiring a separately curated clean reference set, while difficulty-based ranking can assign it high priority.

\newcommand{\figNoise}[2]{%
\begin{figure}[#1]
  \centering
  \includegraphics[width=#2\linewidth]{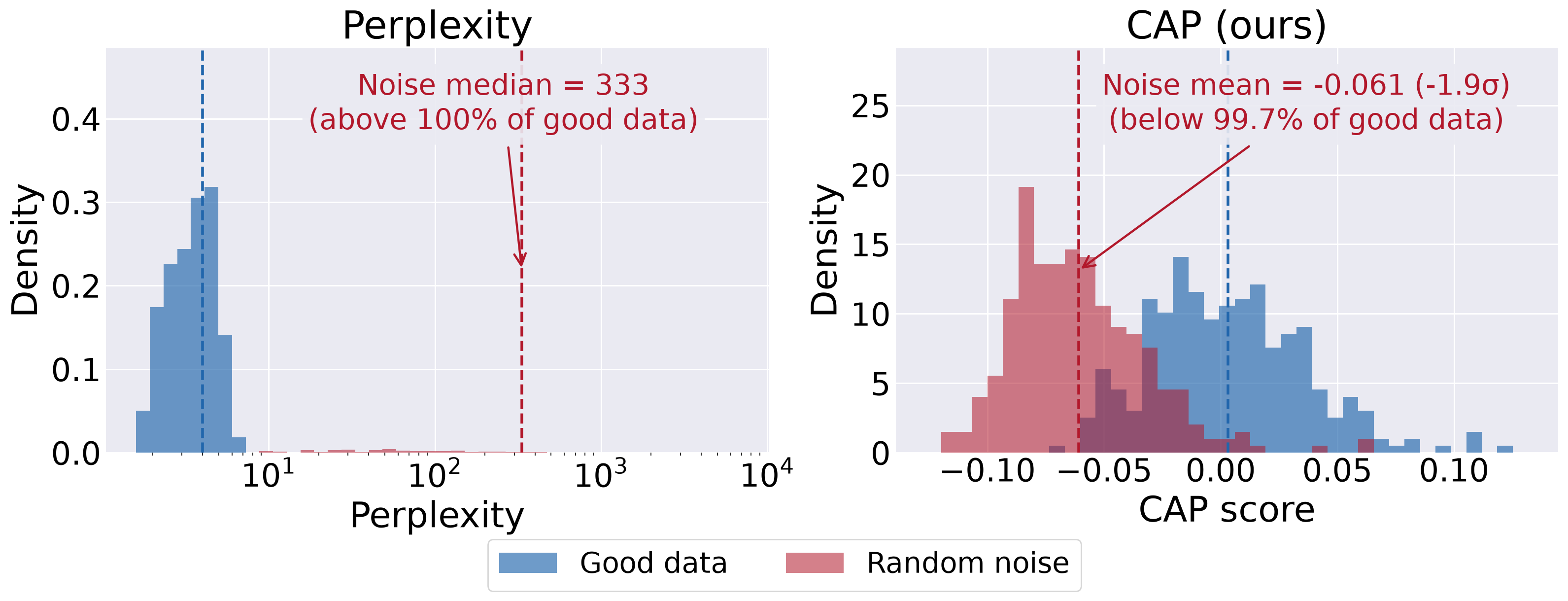}
  \caption{CAP separates unassimilated information from unstructured difficulty. We inject 5\% random junk into a clean pool and score 300 clean and 300 noise examples with Marin-8B-Base. Left: perplexity ranks the noise above the clean distribution. Right: CAP ranks the same noise below 99.7\% of clean examples.}
  \label{fig:noise}
\end{figure}}
\iclronly{\figNoise{ht}{0.95}}
\arxivonly{\figNoise{ht}{0.85}}

\arxivonly{\needspace{20\baselineskip}}
\subsection{The Optimization Path}
\label{sec:optpath}


\begin{wrapfigure}{r}{0.44\linewidth}
  \vspace{-1.1\baselineskip}
  \centering
  \includegraphics[width=\linewidth]{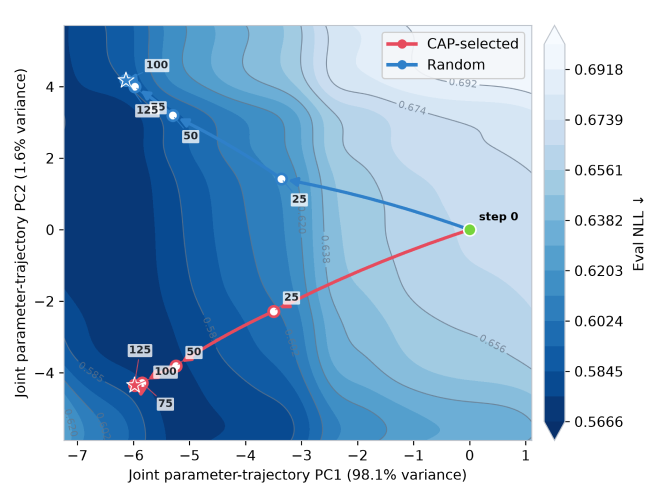}
  \caption{Optimization trajectories under CAP and random selection. Joint-trajectory PCA projection over a shared held-out eval loss surface. The full figure, including held-out loss versus optimizer step, is in Appendix~\ref{app:optpath}.}
  \label{fig:optpath_main}
  \vspace{-1.0\baselineskip}
\end{wrapfigure}

Selection can improve performance either by changing the final solution or by changing the route taken during optimization. We project the 8k math SFT trajectories of CAP and random selection onto the top two principal components of their joint parameter path and evaluate them over a shared held-out GSM8K loss surface. In this projection, the two runs converge to distinct regions rather than different points along the same trajectory. CAP descends more quickly and reaches a lower final held-out loss ($0.572$ versus $0.584$), while random selection begins to degrade late in training (Figure~\ref{fig:optpath_main}; full version in Appendix~\ref{app:optpath}). The difference appears early in training and persists through convergence, suggesting that CAP changes the optimization signal rather than merely improving the final subset average. While the two-dimensional projection cannot establish distinct basins in the full parameter space, it provides complementary evidence that model-aware selection affects not only final performance but also the training dynamics that lead to it.



\section{Conclusion}
\label{sec:conclusion}

We identify an in-and-out information interplay in which weight and hidden-state ranks form complementary profiles across layers, and use this structure to define the \emph{assimilation gap}: information the model has not yet made accessible. This motivates CAP, a lightweight model-aware selection score based on one counterfactual generation and two forward passes. Across math, code, and science, CAP delivers 35.4\% greater average improvement than the strongest baseline, surpasses full-pool training on math with 10\% of the data, and transfers to multimodal and specialized domains. CAP also remains robust across response horizons and rejects unstructured noise that difficulty-based selection can prioritize. Overall, our results suggest that effective post-training data should target what the model has not yet assimilated, rather than what is merely difficult, providing a model-aware view of data value for post-training.\markmainend

\bibliographystyle{plainnat}
\bibliography{references}

\newpage
\appendix

\section{Formal Model, Assumptions, and Proofs}
\label{app:theory}

\subsection{The Memory-Augmented Computation Model}
\label{app:model}

\begin{definition}[Memory-augmented layered computation]
\label{def:model}
A computation of depth $L$ consists of activation capacities
$r_0, \dots, r_L \ge 0$, weight capacities $w_1, \dots, w_L \ge 0$, and a
retrieval schedule $k_1, \dots, k_L \ge 0$ with $\sum_\ell k_\ell = K$, where
$0 \le K \le R$ is the total knowledge retrievable from memory for the task.
Write $K_{>\ell} = \sum_{j>\ell} k_j$. The task requires delivering $R$ units of
predictive information at the output from an input carrying $H \ge cR$ units. The
parameters are a content rate $c > 0$, an addressing function $\varphi$, a
storage rate $\sigma > 0$, a transport rate $\tau \ge 0$, a matching rate
$\sigma_m > 0$, a transport bandwidth $B_t > 0$, and a retrieval bandwidth
$B_r > 0$. A computation is \emph{feasible} if
\begin{itemize}
  \item[(C1)] $r_0 = H$ and $r_L \ge cR$ \hfill (boundary);
  \item[(C2)] $|r_\ell - r_{\ell-1}| \le B_t$ for all $\ell$ \hfill (transport bandwidth);
  \item[(C3)] $0 \le k_\ell \le B_r$ for all $\ell$ \hfill (retrieval bandwidth);
  \item[(C4)] $r_\ell \ \ge\ g(K_{>\ell}) \ :=\ c\,(R - K_{>\ell}) + \varphi(K_{>\ell})$ for all $\ell$ \hfill (information budget);
  \item[(C5)] $w_\ell \ \ge\ \underbrace{\sigma k_\ell}_{\text{injection}}
              \;+\; \underbrace{\tau\,|r_\ell - r_{\ell-1}|}_{\text{transport}}
              \;+\; \underbrace{m_\ell}_{\text{matching}}$, where
              $m_\ell := \sigma_m\,\mu(K_{>\ell})\,\nu(H - r_\ell)$ \hfill (weight cost).
\end{itemize}
An \emph{optimal} computation minimizes $\sum_{\ell=1}^L r_\ell$ over all feasible
$(r,k,w)$ and, among the minimizers, minimizes $\sum_{\ell=1}^L w_\ell$.
\end{definition}

The budget (C4) is the heart of the model: at depth $\ell$ the activations must
carry the content that no later layer will inject, $c(R - K_{>\ell})$, plus an
address for the content that later layers will inject, $\varphi(K_{>\ell})$. The
primary objective encodes the compression bias of training, namely that
representations are as small as the constraints allow; the weight tiebreaker
encodes parameter economy, which is second-order and is used only to select a
minimal $w$ once $(r,k)$ is fixed.

\begin{assumption}[Addressing gain]
\label{ass:addr}
$\varphi : [0, K] \to \mathbb{R}_{\ge 0}$ is concave, non-decreasing,
differentiable, $\varphi(0) = 0$, and $\varphi'(0) \le c - \eta$ for some
$\eta > 0$.
\end{assumption}

Assumption~\ref{ass:addr} states that pointing at stored content is strictly
cheaper than carrying it, which is the defining property of a memory. Concavity
gives $\varphi'(x) \le \varphi'(0) < c$ everywhere, so $g$ in (C4) is strictly
decreasing with slope at most $-\eta$: every unit of retrievable knowledge lowers
the information the activations must carry. Since $K \le R$ we also have
$g \ge 0$ throughout.

\begin{assumption}[Sufficient depth]
\label{ass:depth}
$L \ \ge\ \big(H - g(K)\big)/B_t + \lceil K/B_r \rceil + 1$ and $c\,B_r \le B_t$.
\end{assumption}

\begin{assumption}[Localized matching cost]
\label{ass:match}
$\mu, \nu : \mathbb{R}_{\ge 0} \to \mathbb{R}_{\ge 0}$ are non-decreasing with
$\mu(0) = \nu(0) = 0$, and at least one of them is strictly increasing.
\end{assumption}

The reading is that converting a compressed activation into an address requires
the layer to supply the structure that compression discarded: the cost grows with
the number of undelivered items to be discriminated, $K_{>\ell}$, and with how far
the activation has been compressed below the raw input, $H - r_\ell$. This is the
substantive modeling input behind the weight profile. Nothing below depends on
the functional form beyond monotonicity; we use the bilinear case
$\mu(K') = K'$, $\nu(u) = u/H$ only to state a closed-form ratio.

\begin{assumption}[Matching-dominated regime]
\label{ass:regime}
$\sigma_m\,\mu(K)\,\nu(H - r^\star) \;>\; \big(\sigma + \tau c\big) B_r$, where
$r^\star = g(K)$.
\end{assumption}

Assumption~\ref{ass:regime} says that indexing the store costs more per layer than
reading one bandwidth-unit out of it. It is the condition under which the model
predicts the interior-peaked weight profile of Figure~\ref{fig:ushape}. When it
fails, the model predicts the opposite, a weight profile peaking at the readout
boundary, which makes the prediction falsifiable rather than automatic.

\subsection{The Semantic Factor Model}
\label{app:factor}

\begin{assumption}[Semantic factor model]
\label{ass:factor}
Over the corpus, $(S, G, \varepsilon_e, \varepsilon_l)$ have finite second moments
with $\mathrm{Var}(S) = \sigma_S^2 > 0$, and:
\begin{itemize}
  \item[(F1)] \emph{Exclusion.} $G$ does not load on $\Dearly$. This is inherited
  from Theorem~\ref{thm:shape}(i): the shallow block lies strictly before the
  first retrieval layer $t_r = L - \lceil K/B_r\rceil$, where the memory channel
  has injected nothing.
  \item[(F2)] \emph{Linearity.}
  $\Dearly = a\,S + \varepsilon_e$ and $\Dlate = a'S + b\,G + \varepsilon_l$ with
  $a, a', b > 0$.
  \item[(F3)] \emph{Noise.} $\varepsilon_e, \varepsilon_l$ are mean-zero and
  uncorrelated with each other and with $(S, G)$.
  \item[(N)] \emph{Contaminant homogeneity} (used only in
  Corollary~\ref{cor:noise}). Contaminating examples obey (F2) with the same
  loadings $a, a'$ and with $G = 0$.
\end{itemize}
Define the surface-gap coupling $\gamma = \mathrm{Cov}(S,G)/\sigma_S^2$, the
surface-orthogonal gap $G_\perp = (G - \mathbb{E}G) - \gamma\,(S - \mathbb{E}S)$,
and the attenuation
$\delta = \sigma_{\varepsilon_e}^2/(a^2\sigma_S^2 + \sigma_{\varepsilon_e}^2) \in [0,1)$.
\end{assumption}

Only (F1) is supplied by the capacity model; (F2), (F3) and (N) are modeling
assumptions about the corpus and the measurement. Note also that $\gamma$ and
$\delta$ are not identified from $(\Dearly, \Dlate)$ alone, since (F2) gives a
single equation in the latent $G$ and the loadings $a'$, $b$. The sufficient
conditions of Corollary~\ref{cor:noise} are therefore not directly measurable;
Section~\ref{sec:noise} tests its observable implication instead.

\begin{proposition}[Selection consistency]
\label{prop:select}
Assume in addition that $S$ is independent of $G_\perp$, and that
$\nu_i := \CAP_i - b\,G_{\perp,i}$ are independent across examples and
sub-Gaussian with proxy $\varsigma^2$ conditionally on the gaps. Then for any two
examples with $G_{\perp,i} > G_{\perp,j} + \Delta$,
\[
\Pr\big[\CAP_i < \CAP_j\big] \;\le\; \exp\!\left(-\,\frac{b^2 \Delta^2}{4\varsigma^2}\right),
\]
and top-$m$ selection by CAP retains every example whose gap exceeds the
$m$-th largest gap in the pool by margin $\Delta$, except with probability at
most $n\,e^{-b^2\Delta^2/(4\varsigma^2)}$ over a pool of size $n$.
\end{proposition}

The independence of $S$ and $G_\perp$ is strictly stronger than the
uncorrelatedness that defines $G_\perp$, and is needed because $\nu$ contains the
residual surface term $\delta(a'+b\gamma)(S - \mathbb{E}S)$ of
Theorem~\ref{thm:ident}. It holds automatically if $(S,G)$ is jointly Gaussian.

\subsection{Mapping the Model to Transformers}
\label{app:mapping}

Three modeling choices connect Definition~\ref{def:model} to the measurements of
Section~\ref{sec:obs}.

(M1) \emph{Activation capacity is measured by hidden-state stable rank.} For
representations with approximately isotropic per-direction energy, log-volume, and
hence transmissible information, scales with the number of directions carrying
non-negligible energy, of which $\srk$ is the robust surrogate. The
approximation degrades for strongly anisotropic spectra, which is one reason we
read the profiles qualitatively.

(M2) \emph{Weight capacity is measured by weight stable rank}, read as the number
of directions a layer can read from or write to. This is a coarser proxy than
(M1), since $\srk(W)$ conflates the injection, transport and matching components
of $w_\ell$ into one number; it is consistent with the empirical observation that
weight profiles are noisier than activation profiles.

(M3) \emph{The compression objective is the implicit bias of training}, namely the
information-bottleneck tendency of deep representations combined with the low-rank
bias of gradient descent \citep{huh2023lowrank}.

(M4) \emph{The factor model} takes $D_\ell$ as a capacity-weighted
representational divergence and the depth split $0.33L / 0.66L$ as a conservative
estimate of the phase boundaries $\ell_d$ and $t_r$ of
Theorem~\ref{thm:shape}.

\subsection{Proof of Theorem~\ref{thm:shape}}

\paragraph{Step 0: reduction to $(r,k)$.}
The weight variables appear only in (C5), as a lower bound, and only in the
tiebreaker. Hence for any $(r,k)$ satisfying (C1)--(C4) there is a feasible $w$,
namely (C5) with equality, and the lexicographic optimum is obtained by first
solving for $(r,k)$ and then setting $w$ to that value. It therefore suffices to
minimize $F(r,k) = \sum_{\ell=1}^L r_\ell$ subject to (C1)--(C4).

\paragraph{Existence.}
The feasible set is nonempty (the profile of Step 2 is feasible under
Assumption~\ref{ass:depth}), closed, and restricting to $r_\ell \le H$ is without
loss of optimality, so the feasible set may be taken compact; the objective is
continuous, so a minimizer exists.

\paragraph{Step 1: pointwise lower bound for a fixed schedule.}
Fix $k$. By (C1) and induction on (C2), $r_\ell \ge H - B_t\ell$. Combining with
(C4),
\[
r_\ell \;\ge\; \underline r_\ell(k) \;:=\; \max\{\,H - B_t\ell,\ g(K_{>\ell})\,\} .
\]

\paragraph{Step 2: the bound is attained.}
Set $r_\ell = \underline r_\ell(k)$. Then (C4) holds by construction, and (C1)
holds since $\underline r_0 = \max\{H, g(K)\} = H$ (because
$g(K) \le g(0) = cR \le H$) and $\underline r_L \ge g(K_{>L}) = g(0) = cR$. For
(C2), note that for any two sequences $f, h$,
\[
\big|\max(f,h)_\ell - \max(f,h)_{\ell-1}\big|
\;\le\; \max\big\{|f_\ell - f_{\ell-1}|,\ |h_\ell - h_{\ell-1}|\big\}.
\]
The descent sequence has increments of size exactly $B_t$, and on the $g$
sequence, by the mean value theorem and Assumption~\ref{ass:addr},
\[
g(K_{>\ell-1}) - g(K_{>\ell}) = \big(c - \varphi'(\xi_\ell)\big)k_\ell \in [0,\, c\,k_\ell]
\subseteq [0,\, cB_r] \subseteq [0,\, B_t],
\]
using (C3) and Assumption~\ref{ass:depth}. Hence $\underline r(k)$ is feasible,
and by Step 1 it is the unique optimal profile for the schedule $k$.

\paragraph{Step 3: the optimal schedule is uniquely the maximally deferred one.}
It remains to minimize $F(k) := \sum_\ell \underline r_\ell(k)$. For any feasible
$k$ and any $\ell$, (C3) gives $K_{>\ell} = \sum_{j>\ell} k_j \le (L-\ell)B_r$, and
trivially $K_{>\ell} \le K$, so
\[
K_{>\ell} \;\le\; K^\star_{>\ell} \;:=\; \min\{K,\ (L-\ell)B_r\}.
\]
The bound is attained pointwise by the schedule
$k^\star_\ell = K^\star_{>\ell-1} - K^\star_{>\ell}$, which satisfies
$0 \le k^\star_\ell \le B_r$ and telescopes to
$K^\star_{>0} - K^\star_{>L} = K$ under Assumption~\ref{ass:depth}. Since $g$ is
strictly decreasing, $\underline r(k^\star) \le \underline r(k)$ pointwise for
every feasible $k$, so $k^\star$ is optimal.

For uniqueness, write $\ell_d := \lceil (H - r^\star)/B_t \rceil$, so that
Assumption~\ref{ass:depth} gives $\ell_d \le L - \lceil K/B_r\rceil$. Let $k$ be
optimal. For $\ell \ge \ell_d$ we have
$H - B_t\ell \le r^\star = g(K) \le g(K^\star_{>\ell}) \le g(K_{>\ell})$, so both
$\underline r_\ell(k)$ and $\underline r_\ell(k^\star)$ lie on the $g$ branch; if
$K_{>\ell} < K^\star_{>\ell}$ for some such $\ell$, then
$\underline r_\ell(k) > \underline r_\ell(k^\star)$ while
$\underline r(k^\star) \le \underline r(k)$ everywhere else, giving
$F(k^\star) < F(k)$ and contradicting optimality. Hence
$K_{>\ell} = K^\star_{>\ell}$ for all $\ell \ge \ell_d$. For $\ell < \ell_d$ we
have $K^\star_{>\ell} = K$ and
$K \ge K_{>\ell} \ge K_{>\ell_d} = K^\star_{>\ell_d} = K$, so equality holds there
too. Since $k$ is determined by $\ell \mapsto K_{>\ell}$, we get $k = k^\star$,
which is (i).

\paragraph{Step 4: the activation profile.}
Substituting $K^\star_{>\ell}$ into $\underline r(k^\star)$ gives (ii): a descent
at rate $B_t$ from $H$ to the trough $r^\star = g(K) = c(R-K) + \varphi(K)$
reached at $\ell_d$, a flat segment on $[\ell_d,\, \ell_r]$ with
$\ell_r := L - K/B_r$, and an ascent along $g(K^\star_{>\ell})$ to $g(0) = cR$.
The segment $[\ell_d, \ell_r]$ is nonempty by Assumption~\ref{ass:depth}. The
profile is quasi-convex as the pointwise maximum of a non-increasing and a
non-decreasing function of $\ell$.

\paragraph{Step 5: the weight profile.}
By Step 0, $w_\ell$ equals the right-hand side of (C5) at the optimum. Write
$m_\ell = \sigma_m\,\mu(K^\star_{>\ell})\,\nu(H - r_\ell)$. On the descent,
$k_\ell = 0$, $|r_\ell - r_{\ell-1}| = B_t$, $K^\star_{>\ell} = K$ and
$H - r_\ell = B_t\ell$, so
\[
w_\ell = \tau B_t + \sigma_m\,\mu(K)\,\nu(B_t \ell), \qquad 1 \le \ell \le \ell_d,
\]
with $m_\ell$ non-decreasing in $\ell$. On the trough, $k_\ell = 0$ and
$r_\ell - r_{\ell-1} = 0$, so
\[
w_\ell = m_\ell = \sigma_m\,\mu(K)\,\nu(H - r^\star) =: M,
\qquad \ell_d < \ell \le \ell_r,
\]
constant and equal to the largest value $m$ takes on the descent. On the
retrieval arm, $k_\ell = B_r$ and
$r_\ell - r_{\ell-1} = (c - \varphi'(\xi_\ell))B_r > 0$, so
\[
w_\ell = \sigma B_r + \tau\big(c - \varphi'(\xi_\ell)\big) B_r
       + \sigma_m\,\mu(K^\star_{>\ell})\,\nu(H - r_\ell),
\qquad \ell_r < \ell \le L,
\]
and here $K^\star_{>\ell}$ is strictly decreasing while $r_\ell$ is strictly
increasing, so $m_\ell$ is strictly decreasing, reaching
$m_L = \sigma_m\mu(0)\nu(H - cR) = 0$. Together with $m_0 = \sigma_m\mu(K)\nu(0) = 0$
this gives the quasi-concavity of $m$ with $\arg\max = [\ell_d, \ell_r]$, and
$M$ is strictly increasing in $K$ because $\mu$ is non-decreasing, $\nu$ is
non-decreasing, at least one strictly, and $r^\star(K)$ is strictly decreasing by
Assumption~\ref{ass:addr}.

For the full profile, the boundary values are
$w_1 = \tau B_t + \sigma_m\mu(K)\nu(B_t)$ and
$w_L = \sigma B_r + \tau(c - \varphi'(\xi_L))B_r \le (\sigma + \tau c)B_r$. Under
Assumption~\ref{ass:regime}, $M > (\sigma + \tau c)B_r \ge w_L$, and $M$ likewise
exceeds every value on the retrieval arm beyond the first, so the maximum of $w$
is attained in the interior. In the bilinear case $\mu(K') = K'$, $\nu(u) = u/H$,
\[
\frac{M}{w_1}
= \frac{\sigma_m K (H - r^\star)/H}{\tau B_t + \sigma_m K B_t/H}
= \frac{\ell_d}{1 + \tau H/(\sigma_m K)},
\qquad \ell_d = \frac{H - r^\star}{B_t},
\]
which exceeds one whenever the compression phase spans more than
$1 + \tau H/(\sigma_m K)$ layers. This is (iii). \hfill$\square$

\begin{remark}
The transport term $\tau B_t$ is paid on the descent and on the retrieval arm but
not on the trough, so $w$ itself is quasi-concave only up to that term; the
matching component alone is exactly quasi-concave. We state (iii) in terms of $m$
for this reason. The distinction is immaterial whenever $\tau B_t \ll M$, which
is implied by Assumption~\ref{ass:regime} together with $B_t \ge cB_r$.
\end{remark}

\subsection{Proof of Theorem~\ref{thm:conservation}}
By (C4) and $\varphi \ge 0$, $r_\ell \ge c(R - K_{>\ell}) = cR - cK_{>\ell}$,
which is the first inequality. For the second, (C5) gives
$w_j \ge \sigma k_j$, so $\sum_{j>\ell} w_j \ge \sigma K_{>\ell}$ and hence
$K_{>\ell} \le \sigma^{-1}\sum_{j>\ell} w_j$; substituting into the first
inequality gives the claim. On the retrieval arm of the optimal computation,
Theorem~\ref{thm:shape}(ii) gives
$r_\ell = g(K_{>\ell}) = c(R - K_{>\ell}) + \varphi(K_{>\ell})$, which rearranges
to the stated identity. \hfill$\square$

\begin{remark}
The weight-side inequality is loose by exactly the transport and matching
components of $\sum_{j>\ell}w_j$. The exact conservation is between activation
capacity and \emph{retrievable knowledge}; the version in $w$ is a corollary,
and should not be read as a one-for-one trade between hidden-state rank and
weight rank.
\end{remark}

\subsection{Proof of Corollary~\ref{cor:emergence}}
Let $\psi(K) = cK - \varphi(K)$. By Assumption~\ref{ass:addr},
$\psi'(K) = c - \varphi'(K) \ge \eta > 0$, so $\psi$ is a strictly increasing
bijection onto its range with $\psi(0) = 0$, and
$cR - r^\star(K) = cR - c(R-K) - \varphi(K) = \psi(K)$, giving the meter and its
inverse. The map is affine exactly when $\varphi$ is affine. Monotone deepening
along a trajectory follows by composition with $K(t)$. At $K = 0$, (C4) reads
$r_\ell \ge cR$ for all $\ell$, so
$\underline r_\ell = \max\{H - B_t\ell,\ cR\}$ is non-increasing in $\ell$: there
is no interior minimum below the output value and no re-expansion, hence no U.
\hfill$\square$

\subsection{Proof of Theorem~\ref{thm:ident}}
Write $\hat S = S - \mathbb{E}S$ and absorb means into $\alpha$. Substituting
$G = \mathbb{E}G + \gamma\hat S + G_\perp$ into (F2),
\[
\Dlate = \mathrm{const} + (a' + b\gamma)\,\hat S + b\,G_\perp + \varepsilon_l ,
\]
with $G_\perp$ uncorrelated with $S$ by construction. By (F3) the population
slope is
\[
\beta = \frac{\mathrm{Cov}(\Dlate, \Dearly)}{\mathrm{Var}(\Dearly)}
= \frac{a\,(a' + b\gamma)\,\sigma_S^2}{a^2\sigma_S^2 + \sigma_{\varepsilon_e}^2}
= (1 - \delta)\,\frac{a' + b\gamma}{a} .
\]
The residual is therefore
\[
\CAP = (a' + b\gamma)\hat S + bG_\perp + \varepsilon_l - \beta(a\hat S + \varepsilon_e)
     = \delta(a' + b\gamma)\hat S + bG_\perp + \varepsilon_l - \beta\varepsilon_e ,
\]
which is the stated identity. Claims (i)--(iii) follow by taking covariances
against $G_\perp$ and $S$ and using (F3). \hfill$\square$

\subsection{Proof of Corollary~\ref{cor:noise}}
By (N) a contaminant satisfies (F2) with $G = 0$, so
$G_\perp = -\mathbb{E}G - \gamma(s - \mathbb{E}S)$. Substituting into the identity
of Theorem~\ref{thm:ident} and taking expectations over the noise terms,
\begin{align*}
\mathbb{E}[\CAP \mid S = s, G = 0]
&= \delta(a' + b\gamma)(s - \mathbb{E}S)
   - b\gamma(s - \mathbb{E}S) - b\,\mathbb{E}G \\
&= \big[\delta a' - (1-\delta)b\gamma\big](s - \mathbb{E}S)
   - b\,\mathbb{E}G .
\end{align*}
The bracket is negative iff $\delta(a' + b\gamma) < b\gamma$, i.e.\ iff
$\delta < b\gamma/(a' + b\gamma)$, and the expression is affine in $s$ with that
slope. Any score that is a monotone increasing function of $s$ or of absolute
loss ranks the same points in its upper tail by definition. \hfill$\square$

\begin{remark}
The prediction is affine in the latent surface value $s$. Regressing the observed
$\CAP$ on the observed $\Dearly$ within a contaminant cluster estimates the slope
$\big[\delta a' - (1-\delta)b\gamma\big]\,(1 - \delta_{\mathrm{noise}})/a$, where
$\delta_{\mathrm{noise}}$ is the attenuation within the cluster; only the sign and
the affine shape transfer to measurables without further assumptions.
\end{remark}

\subsection{Proof of Proposition~\ref{prop:select}}
$\CAP_i - \CAP_j = b(G_{\perp,i} - G_{\perp,j}) + (\nu_i - \nu_j)$, and by the
stated independence $\nu_i - \nu_j$ is sub-Gaussian with proxy $2\varsigma^2$.
Thus
\[
\Pr[\CAP_i < \CAP_j] \le \Pr[\nu_j - \nu_i > b\Delta] \le \exp\!\left(-\frac{b^2\Delta^2}{4\varsigma^2}\right),
\]
and the top-$m$ claim follows from a union bound over the at most $n$ comparisons
formed with the threshold example. \hfill$\square$

\subsection{Scope of the Theory}
\label{app:scope}

This section expands the scope discussion of Section~\ref{sec:theory}. Because
the argument spans a capacity model, a mapping to measurements, and a factor
model, it is worth stating in full what each step does and does not carry.

The model is an accounting model, not a model of transformer computation. It
predicts the \emph{shape} of capacity profiles over depth, not their values, and
it is silent about attention, tokenization, and everything else that determines
the constants. Comparisons between predicted and measured curves should be read
qualitatively.

Three claims have different strengths. The activation U and the deferred
retrieval schedule (Theorem~\ref{thm:shape}(i)--(ii)) follow from the addressing
gain assumption alone, which is close to a definition of what a memory is. The
weight inverted U (Theorem~\ref{thm:shape}(iii)) additionally requires
Assumption~\ref{ass:match} on the cost of matching, and holds only in the regime
of Assumption~\ref{ass:regime}; it is the weakest link on the phenomenon side.
The conservation law (Theorem~\ref{thm:conservation}) is exact in $K$ and only an
inequality in $w$: $\sum_{j>\ell} w_j$ also pays for transport and matching, so
the exact trade is between activations and retrievable knowledge, not between
hidden-state rank and weight rank.

Stable rank is a surrogate for capacity, not capacity. For activations this is
the standard argument, exact only when the per-direction energy is approximately
isotropic. For weight matrices we read $\srk(W)$ as the number of directions the
layer can read from or write to, which is a coarser proxy; the empirical weight
profiles are correspondingly noisier than the activation profiles
(Section~\ref{sec:obs}).

The bridge to CAP is an exclusion restriction plus a factor model.
Theorem~\ref{thm:shape} supplies the first, and Theorem~\ref{thm:ident} is an
identification result conditional on the second, not evidence for the factor
model. It nevertheless carries a testable implication that does not depend on the
factor model being right: on a decoder without the rank equilibrium, at
initialization for instance, the shallow block no longer precedes retrieval, the
exclusion restriction fails, and CAP should carry no selection signal. We state
this as a prediction rather than a result, since testing it requires scoring,
fine-tuning, and evaluating with a randomly initialized decoder. Likewise, the
sharpest form of Corollary~\ref{cor:noise} is stated in terms of the latent
surface value $s$; regressing observed $\CAP$ on observed $\Dearly$ within a
noise cluster attenuates the slope by a further factor, so only the sign and the
affine shape are cleanly predicted at the level of measurables. The empirical
case for CAP rests on Section~\ref{sec:main_results}, not on the theory alone.

Finally, the theory predicts a sign and a mechanism, not effect sizes. It does
not determine the depth split fractions, the response horizon $R$, or how much
accuracy any of this should buy.

\section{Full Cross-Model Rank Survey}
\label{app:survey}

\begin{figure}[p]
  \centering
  \includegraphics[width=\linewidth,height=0.78\textheight,keepaspectratio]{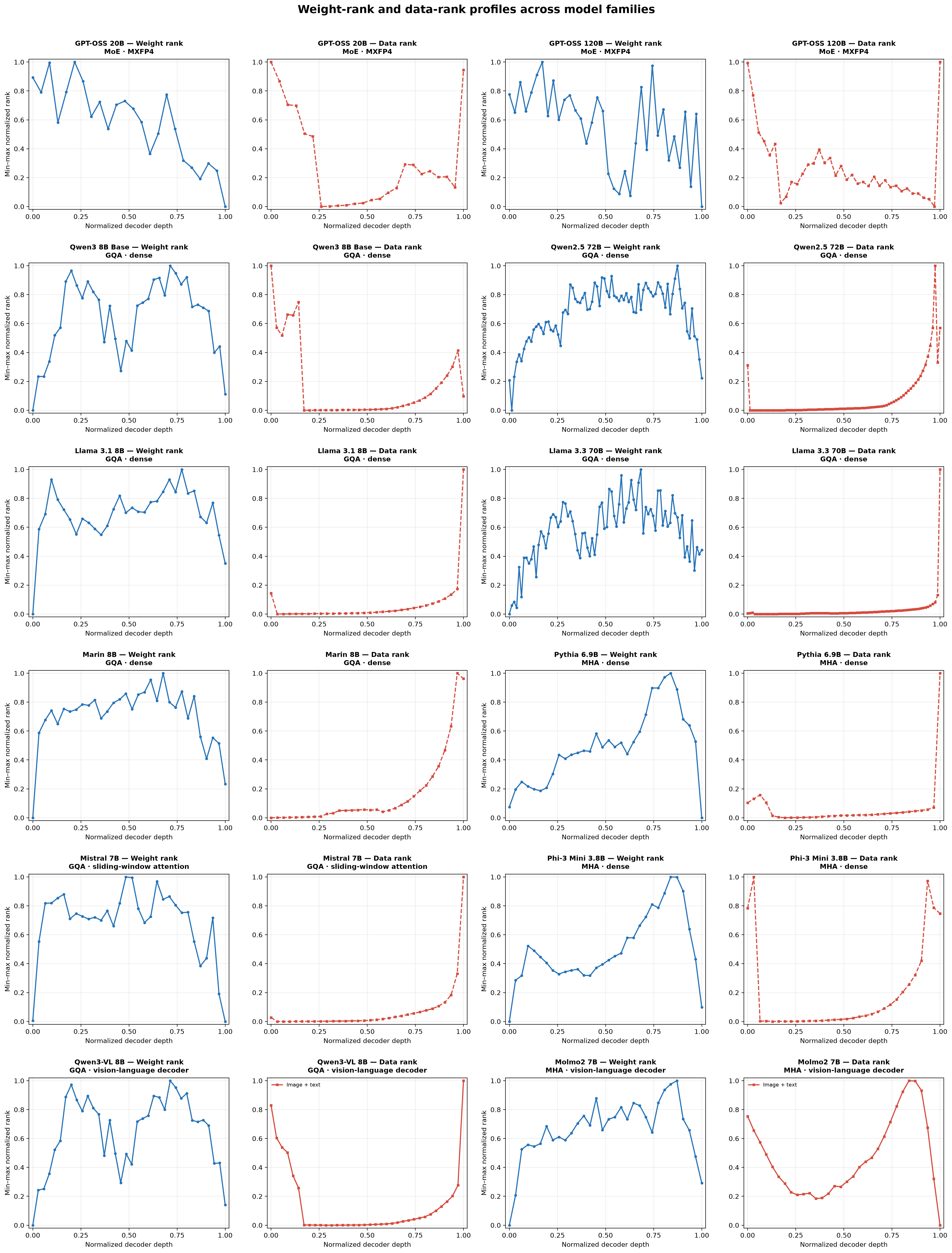}
  \caption{Weight and data stable rank over normalized decoder depth for 12 models. Table~\ref{tab:survey_models} lists their sizes, attention mechanisms, and architectures.}
  \label{fig:rank-survey}
\end{figure}

Figure~\ref{fig:rank-survey} extends the four-model comparison of Figure~\ref{fig:rank4} to 12 models.
Table~\ref{tab:survey_models} describes the checkpoints in the survey.
Parameter counts follow the nominal model sizes in the released checkpoint
names. For GPT OSS, 20B and 120B denote total parameters, with approximately
3.6B and 5.1B active per token, respectively. GQA denotes grouped-query
attention, MHA denotes multi-head attention, and MoE denotes mixture of experts.
For vision-language models, the attention description refers to the decoder.

\begin{table}[t]
  \centering
  \small
  \setlength{\tabcolsep}{4pt}
  \begin{tabular}{@{}>{\raggedright\arraybackslash}p{0.30\linewidth}p{0.13\linewidth}>{\raggedright\arraybackslash}p{0.26\linewidth}ll@{}}
    \toprule
    Model & Size & Decoder attention & Type & Modality \\
    \midrule
    GPT OSS 20B & 20B & GQA, alternating full/window & MoE & Text \\
    GPT OSS 120B & 120B & GQA, alternating full/window & MoE & Text \\
    Qwen3 8B Base & 8B & GQA & Dense & Text \\
    Qwen2.5 72B & 72B & GQA & Dense & Text \\
    Llama 3.1 8B & 8B & GQA & Dense & Text \\
    Llama 3.3 70B Instruct & 70B & GQA & Dense & Text \\
    Marin 8B Base & 8B & GQA & Dense & Text \\
    Pythia 6.9B & 6.9B & MHA & Dense & Text \\
    Mistral 7B v0.3 & 7B & GQA, sliding window & Dense & Text \\
    Phi 3 Mini 4K Instruct & 3.8B & MHA & Dense & Text \\
    Qwen3 VL 8B Instruct & 8B & GQA & Dense & VL \\
    Molmo2 O 7B & 7B & MHA & Dense & VL \\
    \bottomrule
  \end{tabular}
  \caption{The 12-model rank survey. Size is the nominal parameter count;
  GPT OSS sizes are total rather than active parameters. Full/window denotes
  alternating full and sliding-window attention. VL denotes vision-language.}
  \label{tab:survey_models}
\end{table}

\section{Full Main-Domain Results}
\label{app:full_results}

Tables~\ref{tab:full_math}--\ref{tab:full_science} report the complete numerical
results underlying Table~\ref{tab:main} and Figure~\ref{fig:budget_curves}.

\begin{table}[t]
  \centering
  \small
  \begin{tabular}{lcccc}
    \toprule
    Method & 1k & 2k & 4k & 8k \\
    \midrule
    Base model & \multicolumn{4}{c}{61.11} \\
    \midrule
    Random   & \textbf{63.99 {\scriptsize$\pm$ 1.32}} & \underline{65.88 {\scriptsize$\pm$ 1.31}} & 68.01 {\scriptsize$\pm$ 1.28} & 69.37 {\scriptsize$\pm$ 1.27} \\
    PPL      & 62.02 {\scriptsize$\pm$ 1.34} & 64.97 {\scriptsize$\pm$ 1.31} & 65.88 {\scriptsize$\pm$ 1.31} & 68.92 {\scriptsize$\pm$ 1.27} \\
    IFD      & 61.41 {\scriptsize$\pm$ 1.34} & 62.70 {\scriptsize$\pm$ 1.33} & 65.05 {\scriptsize$\pm$ 1.31} & 66.34 {\scriptsize$\pm$ 1.30} \\
    QuRating & 61.11 {\scriptsize$\pm$ 1.34} & 63.76 {\scriptsize$\pm$ 1.32} & 67.78 {\scriptsize$\pm$ 1.29} & 66.11 {\scriptsize$\pm$ 1.30} \\
    LESS     & 60.27 {\scriptsize$\pm$ 1.35} & 64.29 {\scriptsize$\pm$ 1.32} & \underline{68.16 {\scriptsize$\pm$ 1.28}} & \underline{70.81 {\scriptsize$\pm$ 1.25}} \\
    k-center & \underline{63.31 {\scriptsize$\pm$ 1.33}} & 64.44 {\scriptsize$\pm$ 1.32} & 66.11 {\scriptsize$\pm$ 1.30} & 70.58 {\scriptsize$\pm$ 1.26} \\
    \midrule
    CAP (ours) & 62.32 {\scriptsize$\pm$ 1.33} & \textbf{67.78 {\scriptsize$\pm$ 1.29}} & \textbf{74.30 {\scriptsize$\pm$ 1.20}} & \textbf{76.04 {\scriptsize$\pm$ 1.18}} \\
    \bottomrule
  \end{tabular}
  \caption{Full math results on GSM8K.}
  \label{tab:full_math}
\end{table}

\begin{table}[t]
  \centering
  \small
  \begin{tabular}{lcccc}
    \toprule
    Method & 1k & 2k & 4k & 8k \\
    \midrule
    Base model & \multicolumn{4}{c}{30.49} \\
    \midrule
    Random   & 28.05 {\scriptsize$\pm$ 3.52} & \textbf{32.93 {\scriptsize$\pm$ 3.68}} & \underline{32.32 {\scriptsize$\pm$ 3.66}} & 30.49 {\scriptsize$\pm$ 3.61} \\
    PPL      & \underline{30.49 {\scriptsize$\pm$ 3.61}} & 28.05 {\scriptsize$\pm$ 3.52} & 28.05 {\scriptsize$\pm$ 3.52} & \underline{32.93 {\scriptsize$\pm$ 3.68}} \\
    IFD      & 25.00 {\scriptsize$\pm$ 3.39} & 28.66 {\scriptsize$\pm$ 3.54} & 26.83 {\scriptsize$\pm$ 3.47} & 28.66 {\scriptsize$\pm$ 3.54} \\
    QuRating & 29.27 {\scriptsize$\pm$ 3.56} & 28.66 {\scriptsize$\pm$ 3.54} & 31.71 {\scriptsize$\pm$ 3.64} & \underline{32.93 {\scriptsize$\pm$ 3.68}} \\
    LESS     & \textbf{31.71 {\scriptsize$\pm$ 3.64}} & \underline{32.32 {\scriptsize$\pm$ 3.66}} & \textbf{33.54 {\scriptsize$\pm$ 3.70}} & \textbf{35.37 {\scriptsize$\pm$ 3.74}} \\
    k-center & \underline{30.49 {\scriptsize$\pm$ 3.61}} & 27.44 {\scriptsize$\pm$ 3.49} & 29.27 {\scriptsize$\pm$ 3.56} & 32.32 {\scriptsize$\pm$ 3.66} \\
    \midrule
    CAP (ours) & \textbf{31.71 {\scriptsize$\pm$ 3.64}} & 29.88 {\scriptsize$\pm$ 3.59} & \textbf{33.54 {\scriptsize$\pm$ 3.70}} & \textbf{35.37 {\scriptsize$\pm$ 3.74}} \\
    \bottomrule
  \end{tabular}
  \caption{Full code results on HumanEval.}
  \label{tab:full_code}
\end{table}

\begin{table}[t]
  \centering
  \small
  \begin{tabular}{lcccc}
    \toprule
    Method & 1k & 2k & 4k & 8k \\
    \midrule
    Base model & \multicolumn{4}{c}{32.55} \\
    \midrule
    Random   & 33.89 {\scriptsize$\pm$ 0.75} & 35.50 {\scriptsize$\pm$ 0.76} & \underline{37.57 {\scriptsize$\pm$ 0.76}} & 38.19 {\scriptsize$\pm$ 0.77} \\
    PPL      & \textbf{34.83 {\scriptsize$\pm$ 0.76}} & 34.87 {\scriptsize$\pm$ 0.75} & 36.41 {\scriptsize$\pm$ 0.76} & 37.72 {\scriptsize$\pm$ 0.76} \\
    IFD      & 33.85 {\scriptsize$\pm$ 0.75} & 34.56 {\scriptsize$\pm$ 0.75} & 37.48 {\scriptsize$\pm$ 0.76} & 38.27 {\scriptsize$\pm$ 0.76} \\
    QuRating & 34.14 {\scriptsize$\pm$ 0.75} & 35.06 {\scriptsize$\pm$ 0.75} & \textbf{38.76 {\scriptsize$\pm$ 0.76}} & \underline{38.51 {\scriptsize$\pm$ 0.77}} \\
    LESS     & 34.57 {\scriptsize$\pm$ 0.75} & \underline{36.03 {\scriptsize$\pm$ 0.76}} & 36.59 {\scriptsize$\pm$ 0.76} & 38.39 {\scriptsize$\pm$ 0.76} \\
    k-center & 33.78 {\scriptsize$\pm$ 0.75} & 35.07 {\scriptsize$\pm$ 0.76} & 36.29 {\scriptsize$\pm$ 0.76} & 37.08 {\scriptsize$\pm$ 0.76} \\
    \midrule
    CAP (ours) & \underline{34.66 {\scriptsize$\pm$ 0.75}} & \textbf{36.23 {\scriptsize$\pm$ 0.76}} & 37.26 {\scriptsize$\pm$ 0.76} & \textbf{39.05 {\scriptsize$\pm$ 0.77}} \\
    \bottomrule
  \end{tabular}
  \caption{Full science results on MMLU-Pro-5.}
  \label{tab:full_science}
\end{table}

\section{Selection Cost Details}
\label{app:cost}

Table~\ref{tab:cost_full} reports both resource allocation and measured usage for
scoring each 80k-example domain pool. We treat utilization-weighted GPU-hours as
the primary compute measure because it integrates observed GPU utilization over
wall time instead of charging equally for active and idle allocation.

\begin{table}[th]
  \centering
  \small
  \setlength{\tabcolsep}{6pt}
  \begin{tabular}{llrrrr>{\columncolor{yellow!18}}r}
    \toprule
    Domain & Method & \shortstack{Wall\\min.} & GPUs &
    \shortstack{Allocated\\GPU-h} & \shortstack{Avg. GPU\\util. (\%)} &
    \shortstack{Util.-weighted\\GPU-h} \\
    \midrule
    \multirow{7}{*}{Math}
      & Random   & 0.001 & 0 & 0.000 & 0.0 & 0.000 \\
      & PPL      & 4.476 & 4 & 0.298 & 81.6 & 0.243 \\
      & IFD      & 34.159 & 4 & 2.277 & 93.8 & 2.130 \\
      & QuRating & 11.749 & 4 & 0.783 & 53.2 & 0.415 \\
      & k-center & 1.321 & 1 & 0.022 & 10.1 & 0.009 \\
      & LESS     & 60.173 & 4 & 4.012 & 77.2 & 3.086 \\
      & CAP      & 63.476 & 4 & 4.232 & 66.0 & 2.780 \\
    \midrule
    \multirow{7}{*}{Code}
      & Random   & 0.001 & 0 & 0.000 & 0.0 & 0.000 \\
      & PPL      & 2.926 & 4 & 0.195 & 81.7 & 0.159 \\
      & IFD      & 20.222 & 4 & 1.348 & 78.1 & 1.049 \\
      & QuRating & 2.716 & 4 & 0.181 & 85.0 & 0.154 \\
      & k-center & 1.215 & 1 & 0.020 & 11.3 & 0.009 \\
      & LESS     & 52.862 & 4 & 3.524 & 44.6 & 1.566 \\
      & CAP      & 64.047 & 4 & 4.270 & 62.7 & 2.668 \\
    \midrule
    \multirow{7}{*}{Science}
      & Random   & 0.001 & 0 & 0.000 & 0.0 & 0.000 \\
      & PPL      & 7.769 & 4 & 0.518 & 78.2 & 0.403 \\
      & IFD      & 57.617 & 4 & 3.841 & 94.0 & 3.599 \\
      & QuRating & 7.307 & 4 & 0.487 & 89.2 & 0.432 \\
      & k-center & 1.704 & 1 & 0.028 & 7.6 & 0.009 \\
      & LESS     & 79.019 & 4 & 5.268 & 92.7 & 4.873 \\
      & CAP      & 65.685 & 4 & 4.379 & 68.7 & 2.995 \\
    \bottomrule
  \end{tabular}
  \caption{Selection cost for each method and domain. The highlighted column is
  the primary compute measure used in Figure~\ref{fig:cost} and the main-text
  analysis. Values are rounded for readability.}
  \label{tab:cost_full}
\end{table}

\section{Further Text Domains}
\label{app:domains}

CAP is applied unchanged to two additional domains, with the same protocol as the main experiments and a 2k budget. Table~\ref{tab:finance} reports finance and Table~\ref{tab:medicine} reports medicine.

\begin{table}[h]
  \centering
  \small
  \begin{tabular}{lcccc}
    \toprule
    Method & FinanceBench & MMLU-Pro Bus. & MMLU-Pro Econ. & Macro avg \\
    \midrule
    PPL    & 19.33 & 34.60 & 47.39 & 33.77 \\
    Random & 18.67 & \underline{37.01} & \underline{48.82} & \underline{34.83} \\
    CAP (ours) & \textbf{22.00} & \textbf{37.14} & \textbf{49.05} & \textbf{36.06} \\
    \bottomrule
  \end{tabular}
  \caption{Finance: 2k selected from 80k of Sujet-Finance-Instruct \citep{sujet2024finance}, evaluated on FinanceBench \citep{islam2023financebench} and MMLU-Pro splits.}
  \label{tab:finance}
\end{table}

\begin{table}[h]
  \centering
  \small
  \begin{tabular}{lcc}
    \toprule
    Method & Diagnosis accuracy & Reasoning recall \\
    \midrule
    PPL    & \underline{14.38} & 32.71 \\
    Random & 13.60 & \underline{34.72} \\
    CAP (ours) & \textbf{15.16} & \textbf{35.79} \\
    \bottomrule
  \end{tabular}
  \caption{Medicine: 2k selected from the 13k MedCaseReasoning pool \citep{wu2025medcasereasoning}.}
  \label{tab:medicine}
\end{table}

\section{Scaling the Budget on Code}
\label{app:code_scale}

\begin{figure}[t]
  \centering
  \includegraphics[width=0.55\linewidth]{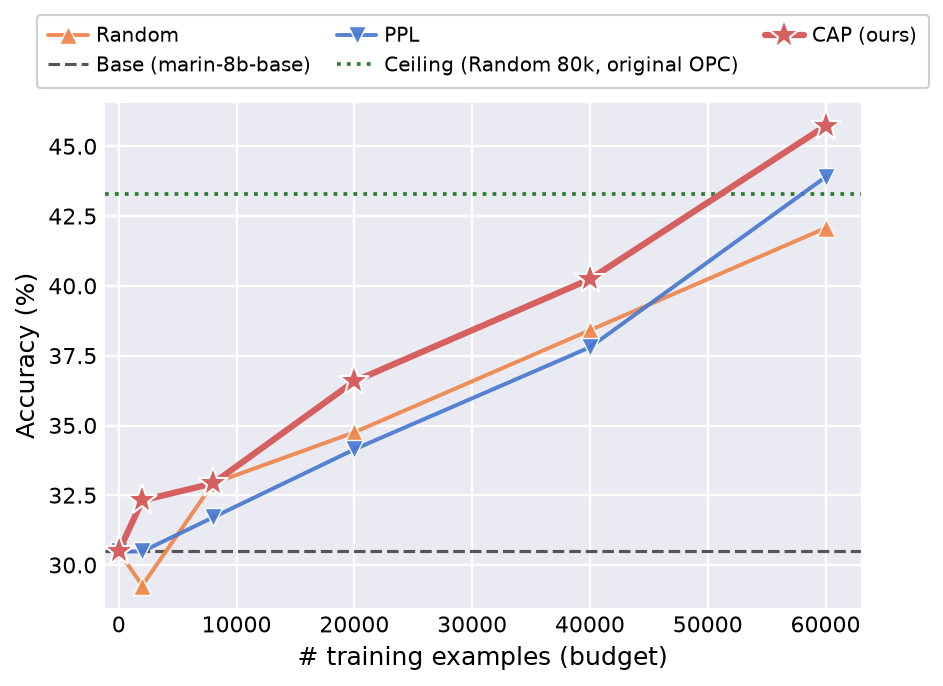}
  \caption{Code accuracy versus enlarged budgets.}
  \label{fig:code_scale_fig}
\end{figure}

Code is the one main-paper domain where a 10\% budget does not match full-pool training (Section~\ref{sec:main_results}), so we extend the sweep to 60k, i.e., 75\% of the pool. Table~\ref{tab:code_scale} and Figure~\ref{fig:code_scale_fig} show that CAP holds a consistent margin over random and PPL at every budget, reaching $+15.24$ over the base model. Selection therefore does not merely reorder the head of the pool: it induces a useful total order over all of it.

\begin{table}[t]
  \centering
  \small
  \begin{tabular}{lcccccc}
    \toprule
    Method & 2k & 8k & 20k & 40k & 60k & Max gain \\
    \midrule
    Random & 29.27 & 32.93 & 34.76 & 38.41 & 42.07 & +11.59 \\
    PPL    & 30.49 & 31.71 & 34.15 & 37.80 & 43.90 & +13.41 \\
    CAP (ours) & \textbf{32.32} & \textbf{32.93} & \textbf{36.59} & \textbf{40.24} & \textbf{45.73} & \textbf{+15.24} \\
    \bottomrule
  \end{tabular}
  \caption{Code at enlarged budgets (HumanEval pass@1, base 30.49).}
  \label{tab:code_scale}
\end{table}

\section{The Optimization Path}
\label{app:optpath}

\begin{figure}[t]
  \centering
  \includegraphics[width=0.95\linewidth]{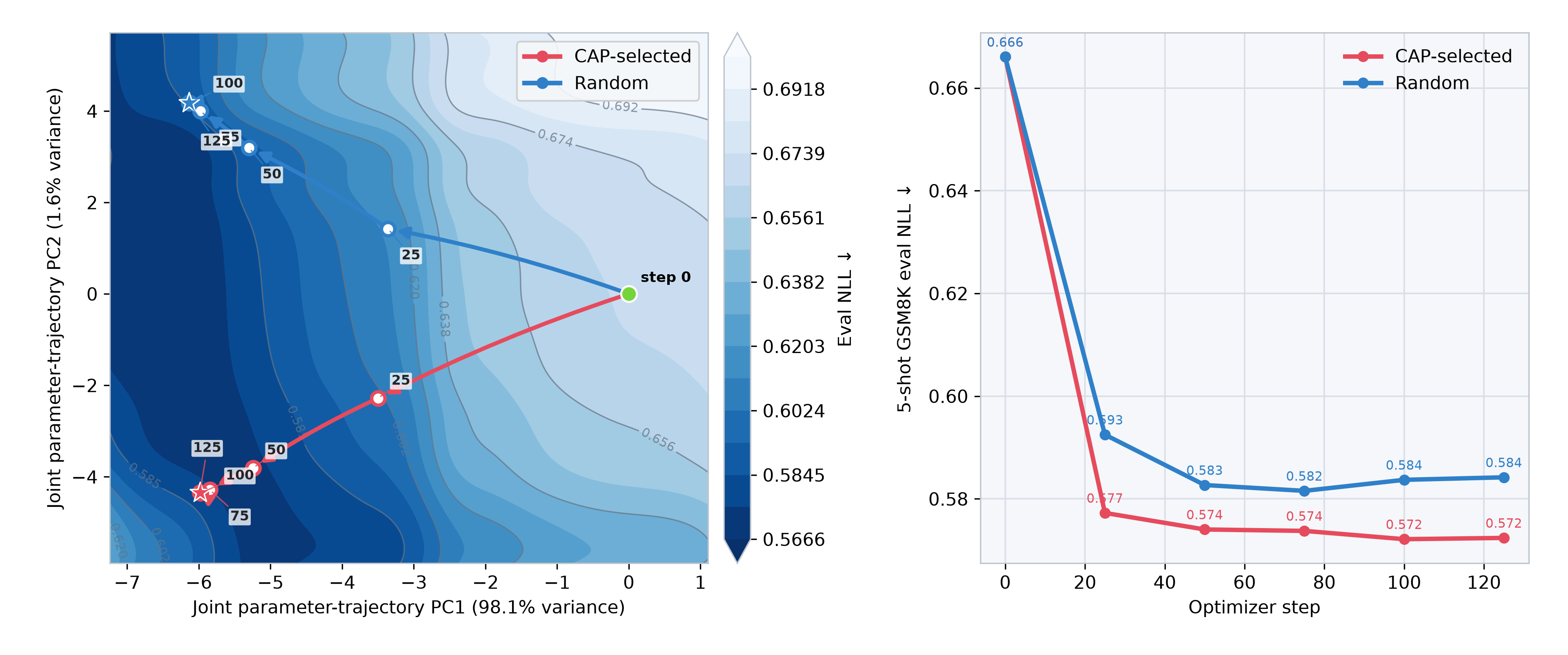}
  \caption{SFT trajectories of CAP and random selection in the plane spanned by the top two PCs of the joint parameter path, over a shared held-out 5-shot GSM8K loss surface.}
  \label{fig:optpath}
\end{figure}

Selection quality can show up in two ways: a better final solution, or a faster route to it. To separate them we take the two 8k math runs that differ only in which examples were selected, save parameters every 25 optimizer steps, and project both paths onto the top two principal components of their joint trajectory, which capture 89\% of its variance. On that plane we evaluate a single shared objective, the negative log-likelihood of held-out 5-shot GSM8K responses, so both runs are measured against the same surface rather than against their own training losses (Figure~\ref{fig:optpath}).

The two paths leave the shared base model, at loss $0.666$, in opposite directions along the second component and never return to each other: the runs end in different basins, not at different points of one basin. CAP descends faster and further. After 25 steps it is already at $0.577$, below the best value random selection reaches at any point of its run ($0.582$ at step 75), and it settles at $0.572$ against $0.584$. Random selection also degrades late, rising from $0.582$ at step 75 to $0.584$ by step 125, whereas the CAP path is monotone to within noise. Both observations are consistent with what the score selects. A set chosen for high assimilation gap poses one coherent demand on the weights, so its per-example gradients plausibly reinforce rather than cancel, and the run walks into a lower basin instead of oscillating inside a shallow one.

\section{Training and Evaluation Details}
\label{app:training}

\subsection{Supervised Fine-Tuning Protocol}

All text experiments fine-tune \texttt{marin-community/marin-8b-base} with full-parameter SFT using LlamaFactory, driven by a single sweep script so that every selection method is trained by an identical pipeline. The only thing that varies between cells of a table is the list of selected example indices; the pool, the budget, the shuffling order, the random seed, the optimizer state, the schedule, and the number of epochs are held fixed. Table~\ref{tab:hparams} lists the configuration.

\begin{table}[t]
  \centering
  \small
  \begin{tabular}{ll}
    \toprule
    Setting & Value \\
    \midrule
    Base model & \texttt{marin-community/marin-8b-base} \\
    Fine-tuning type & full parameter (no adapters) \\
    Chat template & \texttt{llama3} \\
    Max sequence length & 2048 tokens \\
    Sequence packing & disabled \\
    Optimizer & AdamW (\texttt{adamw\_torch}) \\
    Peak learning rate & $1\times 10^{-5}$ \\
    LR schedule & cosine decay \\
    Warmup ratio & 0.03 \\
    Epochs & 1 \\
    Per-device batch size & 2 \\
    Gradient accumulation & 8 \\
    Effective batch size & 64 ($2 \times 8 \times 4$ GPUs) \\
    Precision & bf16 \\
    Gradient checkpointing & enabled \\
    Distributed backend & DeepSpeed ZeRO stage 3 \\
    Hardware & 4$\times$ NVIDIA A100 SXM \\
    \bottomrule
  \end{tabular}
  \caption{SFT configuration, shared by every method and budget.}
  \label{tab:hparams}
\end{table}

Two choices deserve comment because they affect how the results should be read. First, we train for a single epoch at every budget. This means that the number of optimizer steps grows linearly with the budget, so a budget curve conflates ``more data'' with ``more updates''; the comparison between methods at a fixed budget is nevertheless exact, since all methods take the same number of steps on the same schedule. Second, the maximum sequence length of 2048 tokens covers the large majority of responses in all three pools, but it truncates a small tail of long chain-of-thought solutions in the math pool; truncation is applied identically to every method.

\subsection{Pool Construction}

\paragraph{Normalization and sampling.}
We use the same candidate pool for every selection method within a domain. Text-pool deduplication applies Unicode NFKC normalization, Unicode case folding, whitespace collapse, and trimming. The key is the normalized instruction together with the optional input, not the response. We retain the first occurrence of each key and sample without replacement with seed 0. Each of the three main domains contains 80,000 sampled examples.

\paragraph{Main text domains.}
For math, the source is the 859,494-example training split of NuminaMath-CoT \citep{numina2024}. For code, we use the 118,278-example \texttt{educational\_instruct} split of OpenCoder opc-sft-stage2 \citep{huang2024opencoder}; prompt deduplication leaves 92,813 unique examples, from which we sample 80,000. This removes 25,465 duplicates, or $21.53\%$ of the source split. For science, we restrict the 2,335,220-example training split of WebInstructSub \citep{yue2024mammoth2} to its 317,209-example Science StackExchange subset covering physics, biology, chemistry, and computer science, and sample 80,000 examples. Separate post-deduplication counts for the math and science sources are not available in the pool summary, so we do not report duplicate-removal rates for those two domains.

\paragraph{Additional text pools.}
The filtered Sujet-Finance source \citep{sujet2024finance} contains 86,217 valid and unique prompts before sampling 80,000 examples. MedCaseReasoning \citep{wu2025medcasereasoning} contains 13,092 valid and unique prompts. The MedMCQA pool contains 182,822 valid rows and 182,788 unique prompts, removing 34 duplicates, or $0.019\%$ of valid rows. These counts distinguish source filtering and deduplication from the subsequent selection budget.

\paragraph{Multimodal document pool.}
We presample 23,000 document-domain candidates. The source is Molmo2-SynMultiImageQA. Deduplication on normalized questions retains 22,984 candidates, removing 16 duplicates, or $0.070\%$, after which we uniformly sample 20,000 examples. Image bytes are not part of this deduplication key. Consequently, identical normalized questions paired with different documents can be merged by this protocol.

\subsection{Scoring Configuration}

\paragraph{CAP.}
The text scoring pipeline uses the raw tokenizer rather than the SFT chat template. We strip the instruction and, when an input is present, append a newline followed by the stripped input. The reference branch appends the stripped response without adding response special tokens. We retain the last 256 prompt tokens and the first $R=48$ response tokens, excluding examples with fewer than two prompt tokens or six response tokens. The counterfactual branch generates from the same truncated prompt using greedy decoding, with sampling disabled, one beam, and at most 48 new tokens. Both branches run in bfloat16. Hidden states are collected from decoder-block outputs at response positions; only blocks in the first and final thirds are retained. Specifically, $\Dearly$ uses layers up to $0.33L$ and $\Dlate$ uses layers from $0.66L$. The residualizing regression is fitted separately on each domain pool. Examples are sorted by tokenized length and scored in batches of 16. The response-horizon ablation varies $R$ as described in Section~\ref{sec:ablate_R}.

\paragraph{PPL.}
Perplexity scoring uses the same 256-token prompt limit, 48-token response limit, six-token response minimum, bfloat16 model, and batch size of 16 as CAP.

\paragraph{QuRating.}
We score the full example text. The input concatenates the instruction, optional input, and response. The rater is \path{princeton-nlp/QuRater-1.3B}, and the selection signal is the \texttt{educational\_value} logit. The maximum sequence length is 512 tokens and the batch size is 16.

\paragraph{IFD.}
We score one example at a time on each GPU. Four strided scoring shards run concurrently across four GPUs, giving a per-GPU batch size of 1 and an aggregate concurrent batch size of 4. Each example is truncated to a maximum sequence length of 1,024 tokens. The score is the ratio of response perplexity conditioned on the instruction to response perplexity without the instruction. Prompt tokens are masked from the conditional loss, and examples with an IFD ratio greater than or equal to one are excluded from selection.

\paragraph{k-center.}
We embed the concatenated prompt and response. The encoder is \path{sentence-transformers/all-MiniLM-L6-v2}; we mean-pool token representations and normalize each embedding to unit length. The embedding maximum length is 256 tokens and the batch size is 256. Thus the geometric baseline uses both the prompt and response, rather than instruction-only embeddings.

\paragraph{LESS.}
The selection cost comparison uses LESS 8192 without gradient warmup. The target set contains at most 2,000 examples from the downstream domain: the first 2,000 GSM8K training examples for math, all 164 HumanEval problems with their canonical solutions for code, and the complete 285-example MMLU validation split for science, comprising five examples from each of 57 subjects. The science target is drawn from MMLU, not MMLU-Pro. The code target overlaps the entire HumanEval evaluation set and includes its reference solutions; this cost configuration therefore does not use a disjoint target set for code.

LESS attaches rank-8 LoRA modules with scaling factor 16 and zero dropout to the query, key, value, and output projection matrices. For each candidate and target example, it computes the gradient of the response-only causal language modeling loss with respect to all trainable LoRA parameters, using a maximum sequence length of 1,024 tokens. The full LoRA gradient is compressed to an 8,192-dimensional vector using a fixed signed CountSketch projection and then normalized to unit length. Target vectors are averaged and normalized, and each candidate is scored by cosine similarity to this aggregate target gradient.

Candidate and target gradients are computed one example at a time on each GPU, with four strided shards running concurrently across four GPUs. Thus the LESS scoring batch size is 1 per GPU and the aggregate concurrent batch size is 4.

\subsection{Evaluation}

\paragraph{Shared configuration.}
The three main text benchmarks use lm-evaluation-harness version \texttt{0.4.13.dev0}, with bfloat16 inference, a model context length of 4,096 tokens, and evaluation batch size 16.

\paragraph{GSM8K.}
We evaluate all 1,319 test examples with five demonstrations from the training split. Each item is formatted as \texttt{Question: \{question\}} followed by a newline and \texttt{Answer:}; demonstrations are separated by two newlines. Decoding is greedy with temperature 0 and at most 1,024 generated tokens, stopping before a new \texttt{Question:} or an end token. We report strict exact match after extracting the number following \texttt{\#\#\#\#} and removing commas, dollar signs, and a final period.

\paragraph{HumanEval.}
We evaluate all 164 test problems with zero demonstrations, using the canonical function prefix supplied by the benchmark directly. Decoding is greedy with at most 1,024 generated tokens and stops on a new class definition, function definition, comment, \texttt{if}, or \texttt{print} statement. We report pass@1 using the official execution-based tests.

\paragraph{MMLU-Pro STEM.}
The reported score is the macro-average over biology, chemistry, computer science, engineering, and physics. We evaluate the complete test subsets of 717, 1,132, 410, 969, and 1,299 examples, respectively, for a subsampling fraction of 1.0 in every subject. Each subject uses the first five chain-of-thought demonstrations from its validation split. The prompt asks the model to reason step by step and finish with \texttt{the answer is (X)}. Decoding is greedy with temperature 0, at most 512 generated tokens, and \texttt{Question:} as the stop sequence. Answers are extracted using the case-insensitive regular expression
\begin{verbatim}
answer is \(?([ABCDEFGHIJ])\)?
\end{verbatim}
and exact match ignores case and punctuation.

\finishdocument

\end{document}